\documentclass{article}
\usepackage{arxiv}
\usepackage[utf8]{inputenc}
\usepackage[T1]{fontenc}
\usepackage{hyperref}
\usepackage{url}
\usepackage{booktabs}
\usepackage{amsfonts}
\usepackage{nicefrac}
\usepackage{lipsum}
\usepackage{graphicx}
\usepackage[numbers,sort&compress]{natbib}
\usepackage{doi}
\usepackage{float}
\usepackage{makecell}
\usepackage[normalem]{ulem}
\usepackage{cancel}
\usepackage{capt-of}
\usepackage{longtable}
\usepackage{caption}
\usepackage{enumitem}
\usepackage{mathpazo}

\usepackage{pdfpages}

\usepackage{amsmath}
\usepackage{subcaption}
\usepackage{booktabs}
\usepackage[most]{tcolorbox}

\usepackage{CJKutf8}

\usepackage{tcolorbox}
\usepackage{newfloat}
\DeclareFloatingEnvironment{boxes}

\newcommand{\boxref}[1]{\hyperref[{#1}]{\ref*{#1}}}
\usepackage{chngcntr}

\title{BeamPERL: Parameter-Efficient RL with Verifiable Rewards Specializes Compact LLMs for Structured Beam Mechanics Reasoning}

\date{}

\author{
\href{https://orcid.org/0009-0005-6969-1852}{\includegraphics[scale=0.06]{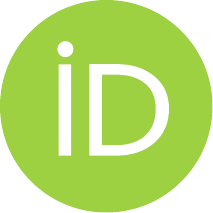}\hspace{1mm}Tarjei Paule Hage} \\
Department of Mechanical Engineering \\
Massachusetts Institute of Technology \\
Cambridge, MA, USA
\And
\href{https://orcid.org/0000-0002-4173-9659}{\includegraphics[scale=0.06]{orcid.pdf}\hspace{1mm}Markus J. Buehler} \\
Department of Civil and Environmental Engineering \\
Department of Mechanical Engineering \\
Schwarzman College of Computing \\
Massachusetts Institute of Technology \\
Cambridge, MA, USA \\
\texttt{mbuehler@MIT.EDU}
}

\makeatletter
\providecommand{\headeright}{}
\providecommand{\undertitle}{}
\providecommand{\shorttitle}{}
\makeatother
\renewcommand{\shorttitle}{BeamPERL: PE-RLVR-FT for Beam Mechanics Problem-Solving}

\hypersetup{
pdftitle={BeamPERL},
pdfsubject={cs.AI},
pdfauthor={Tarjei P.~Hage, Markus J.~Buehler},
pdfkeywords={Large Language Models, Reinforcement Learning, GRPO, Parameter-Efficient Fine-Tuning, Engineering Reasoning, Beam Mechanics},
}

\begin{document}
\maketitle
\begin{abstract}
\label{sec:abstract}
Can reinforcement learning with hard, verifiable rewards teach a compact language model to reason about physics, or does it primarily learn to pattern-match toward correct answers? We study this question by training a 1.5B-parameter reasoning model on beam statics, a classic engineering problem, using parameter-efficient RLVR with binary correctness rewards from symbolic solvers, without teacher-generated reasoning traces. The best BeamPERL checkpoint achieves a 66.7\% improvement in Pass@1 over the base model. However, the learned competence is anisotropic: the model generalizes compositionally (more loads) but fails under topological shifts (moved supports) that require the same equilibrium equations. Intermediate checkpoints yield the strongest reasoning, while continued optimization degrades robustness while maintaining reward. These findings reveal a key limitation of outcome-level alignment: reinforcement learning with exact physics rewards induces procedural solution templates rather than internalization of governing equations. The precision of the reward signal - even when analytically exact - does not by itself guarantee transferable physical reasoning. Our results suggest that verifiable rewards may need to be paired with structured reasoning scaffolding to move beyond template matching toward robust scientific reasoning.

\end{abstract}


\section{Introduction}
\label{sec:introduction}
In a future where engineers and intelligent systems work side by side, artificial intelligence (AI) models are poised to serve as the backbone of agentic and other scientific workflows \cite{vaswani2017attention,Wei2021FinetunedLearners,wei2022chainofthought,10.1063/5.0134317,wang2023scientific,berens2023ai,Buehler2024Accelerating,ghafarollahi_sciagents_2025,stewart2026graphagentsknowledgegraphguidedagentic,Buehler2026PhysicsAwareEmergentCognition,lu_fine-tuning_2025}. These systems will increasingly take on substantial portions of engineering tasks, with human experts validating and refining their outputs to achieve greater efficiency, reliability, and scale in engineering practice \cite{ghafarollahi_sparks_2025}. As a result, rapid progress is occurring on two fronts: the development of massive large language models (LLM) with broad general-purpose abilities and the creation of smaller, specialized LLMs tailored to specific tasks \cite{chen_what_2025}. Although there is not a consensus on which design philosophy will ultimately dominate in future agentic engineering systems, it is clear that compute-efficient model development, training, and deployment will be essential \cite{arora_training_2025}.

LLMs are being applied across the scientific supply chain, from idea generation \cite{ghafarollahi_sciagents_2025} and materials discovery \cite{ghafarollahi_rapid_2025} to structural analysis \cite{liang_integrating_2025} and mechanical design \cite{elrefaie_ai_2025}. Reasoning processes, which mirror how humans conduct scientific inquiry and engineering workflows, play a central role in enabling these capabilities. In response, several early approaches to machine reasoning have emerged~\cite{wei2022chainofthought,zelikman2022star,zelikman2024quietstar,buehler2024xlora}. Agentic systems decompose workflows into discrete steps handled by specialized LLM-based agents that collaborate to achieve a common goal \cite{shi_deep_2025}, while Large Reasoning Models (LRM) perform internal multi-stage reasoning to critique and refine their own outputs before producing a final answer to a given task \cite{zhang_survey_2025}. A combination of these approaches, in which agents in the agentic systems make use of LRMs themselves, has the potential to play a central role in the engineering pipelines of the future. LRMs, in particular, are central to this emerging paradigm because their internal reasoning processes allow models to solve complex engineering problems in a manner that more closely mirrors human analytical practice \cite{li_system_2025}.

LRMs, LLMs that leverage increased test-time compute to produce internal reasoning traces before committing to a final answer, have revolutionized mathematical reasoning in artificial intelligence, with significant advancements across various benchmarks \cite{zhang_survey_2025}. More broadly, the fine-tuning of both LLMs and LRMs has become a central driver of progress, enabling models to improve not only on mathematical and coding tasks but also across a wide range of real-world applications \cite{xu_toward_2025}. Yet, while these general-purpose improvements provide the foundation for scientific and engineering reasoning, they also raise an important question: Can we further enhance these models’ capabilities on specific engineering tasks through task-specific fine-tuning, thereby improving LRM performance within specialized engineering systems?

For specialized engineering applications, it has been shown that good performance can be achieved without massive models or extensive training pipelines \cite{luu_bioinspiredllm_2024}. Engineering domains such as beam mechanics rely on structured physical principles, symbolic manipulation, and verifiable mathematical relationships, which may not require the broad world knowledge captured by massive, general-purpose LLMs. This raises the possibility that compact reasoning models, when adapted with targeted training signals, can learn to navigate these well-defined problem spaces with high accuracy \cite{lu_fine-tuning_2025}. In such domains, the requirements for domain competence may be less dependent on scale and more dependent on precise alignment with the underlying governing equations. This perspective motivates a minimalist approach to engineering-oriented language models. Rather than relying on vast datasets or costly full-parameter training, it may be possible to construct small, task-specific models capable of internalizing the structure of an engineering problem through reinforcement learning (RL) \cite{sutton_reinforcement_2014}. Such models could acquire the essential reasoning patterns needed for a narrow task while avoiding the overhead associated with large-scale general-purpose fine-tuning. If successful, this paradigm offers a path toward lightweight, specialized reasoning agents that deliver reliable performance in engineering workflows without the computational cost typically associated with advanced LRMs.

A key precursor to the present work is PRefLexOR (Preference-based Recursive Language Modeling for Exploratory Optimization of Reasoning)~\cite{buehler_preflexor_2025,Buehler2025GPReFLexOR}, which established that language models can acquire enhanced reasoning capabilities through outcome-level optimization without explicit supervision of intermediate reasoning steps. PRefLexOR introduced a two-phase training strategy: a first phase of structured thought integration using preference optimization that teaches the model to generate reasoning within dedicated thinking tokens, followed by a second phase of independent reasoning development in which intermediate reasoning tokens are masked during training, forcing the model to implicitly discover effective reasoning pathways guided only by alignment with verified final answers. This masking mechanism is central to the present work's motivation: by withholding explicit reasoning traces, PRefLexOR demonstrated that a model can learn how to reason rather than merely imitating predetermined solution paths. Notably, the work showed that even compact models (3B parameters) can self-teach sophisticated scientific reasoning through this approach, including the ability to generalize reasoning strategies to tasks outside the training domain. The earlier work in open-domain scientific reasoning raised a natural follow-up question that the present study directly pursues: can comparable or stronger specialization be achieved when the training signal shifts from preference-derived quality assessments to deterministic, analytically verifiable binary rewards derived from symbolic solvers - and does this harder, more precise reward signal lead to genuine internalization of governing physical equations, or to narrowly scoped procedural specialization?

We hypothesize that parameter-efficient reinforcement learning with verifiable rewards is sufficient to improve dense, distilled language models on standardized mechanics problems without requiring teacher-generated reasoning traces. We further hypothesize that if such alignment induces genuine internalization of governing equations rather than reward-driven pattern specialization, the resulting competence will generalize robustly across variations in problem configuration and maintain reasoning coherence. To test this minimalist paradigm, we construct a controlled baseline experiment in beam mechanics problem-solving, a domain grounded in structured physical principles and verifiable mathematical relationships. We adapt a dense LRM using parameter-efficient fine-tuning (PEFT) through reinforcement learning with verifiable rewards (RLVR) to compute the reaction forces of a statically loaded beam. The model learns to produce structured reasoning traces that yield correct equilibrium calculations by leveraging Group Relative Policy Optimization (GRPO) with both accuracy and format rewards. This setup isolates outcome-level alignment without teacher-generated traces or auxiliary supervision, allowing us to evaluate whether such minimal alignment is sufficient to produce strong in-distribution performance and robust generalization across variations in problem configuration. If successful, this baseline would support the view that language models, given a verifiable objective and limited adaptation, can autonomously acquire domain-specific scientific reasoning. More broadly, it establishes a reference point for assessing whether outcome-level alignment alone yields transferable engineering reasoning or instead induces distribution-dependent specialization.

\subsection{Reinforcement Learning for Incentivizing Reasoning}
\label{subsec:LRM}

Recent advancements in LRMs have been driven largely by RL post-training applied to large pre-trained LLMs. RL-based post-training has gained prominence through methods such as Reinforcement Learning from Human Feedback (RLHF) \cite{christiano_deep_2023}, which aligns model outputs with human preferences. LRMs build on this foundation by using RL not to align behavior, but to directly incentivize and strengthen reasoning \cite{zhang_survey_2025}. This new class of models began with OpenAI’s o1 series \cite{openai_openai_2024}, the first to explicitly introduce inference-time scaling, where large-scale RL is used to extend and refine the model’s chain-of-thought reasoning~\cite{wei2022chainofthought}, leading to substantial improvements in reasoning capability and robustness.

Much of the development of leading reasoning models remains closed source, motivating substantial effort toward open-source training algorithms for LRMs. Earlier work such as PRefLexOR~\cite{buehler_preflexor_2025} demonstrated that outcome-level optimization with masked reasoning tokens can enhance scientific reasoning in compact models without requiring explicit supervision of intermediate steps. PRefLexOR employed preference-based optimization (ORPO and EXO) with reward signals derived from corpus-grounded answer quality (signals that capture relative correctness across a broad scientific domain). A central open question following that work was whether shifting the reward signal toward the harder end of the spectrum — specifically, to deterministic binary rewards encoding strict physical correctness from symbolic solvers — would yield comparable or stronger reasoning improvements, particularly in engineering domains where ground-truth solutions are analytically available.

A pivotal early demonstration of RLVR-style reasoning was the DeepSeek-R1 series \cite{deepseek-ai_deepseek-r1_2025}, which showed that reinforcement learning fine-tuning (RLFT) can substantially improve mathematical reasoning, particularly when combined with a cold-start SFT phase. Subsequent analyses have moved beyond raw benchmark gains to examine the qualitative transformations induced by RL, including verification behaviors and structured exploration, and other reasoning patterns associated with improved performance \cite{kim_reasoning_2026}. Moreover, the authors showed that the reasoning patterns learned by larger RL-trained LRMs can be effectively distilled into smaller models through SFT on extracted reasoning traces. These distilled models outperform other instruction-tuned models built on the same base checkpoints and, notably, exceed the performance of small models trained directly with RL. This highlights a key insight: Reasoning knowledge discovered by large RL-trained models can be transferred to denser models via SFT, enabling compact models to inherit sophisticated reasoning behaviors that could be difficult for them to acquire through RL alone. This observation expands our initial motivating questions to the following: Can RLFT applied to small, distilled LRMs yield measurable improvements in their ability to perform standardized mechanics problems when adapted with task-specific, verifiable training signals?

\subsection{Open-Source Reproductions and Efforts Toward Efficient LRM Development}
\label{subsec:LRM_reproductions}

Following the release of DeepSeek’s R1 model series, there has been effort by the open-source community to reproduce it's training pipeline, which was not released. Among these are Hugging Face’s Open R1 project \cite{hugging_face_open_2025}, a fully open source reproduction of DeepSeek-R1, focused on eliciting reasoning in language models via RLVR. Other projects include STILL-3 \cite{chen_empirical_2025} and DAPO \cite{yu_dapo_2025}, which provide fully open-source frameworks for large-scale GRPO training and introduces algorithm modifications that improve training effectiveness. Furthermore, Oat-Zero \cite{liu_understanding_2025}, another framework, investigates how the pretraining characteristics of the base model influence pure RL training outcomes.

While the DeepSeek-R1 and R1-Zero models are based on DeepSeek-V3-Base, a 671B-parameter base model, they observed that for smaller models (1.5B–70B) pure RL training, i.e., no zero–cold-start SFT phase prior to RLFT, was less effective than distilling reasoning capabilities through SFT on traces generated by the larger R1 model. This observation prompted several investigations within the open-source community into how RLFT performs on dense base models. Open-Reasoner-Zero \cite{hu_open-reasoner-zero_2025} provides an open-source implementation of large-scale zero RL training on dense models, though it employs a PPO-style algorithm \cite{schulman_proximal_2017} rather than GRPO. Complementing this, STILL-3 \cite{chen_empirical_2025} demonstrates that also GRPO-based training can improve the accuracy of smaller models across multiple mathematical reasoning benchmarks. Finally, SimpleRL-Zoo \cite{zeng_simplerl-zoo_2025} conducts pure GRPO RL training across a diverse set of dense base models, offering a systematic view of what conditions enable small LLMs to meaningfully improve through RL alone.

The authors noted that pure RL training of base models, although improving performance on reasoning benchmarks, also introduced challenges such as poor readability and language mixing. To mitigate these issues, they incorporated a cold-start phase into a multi-stage training pipeline, during which the model was SFT on thousands of reasoning traces collected from various sources before undergoing subsequent RL refinement. Not only did this improve the model's coherency, but also enhanced its reasoning performance. Recent work, such as the Open-RS project \cite{dang_reinforcement_2025}, has drawn an explicit analogy to this process for dense models, suggesting that using a distilled reasoning model as the base model plays a role analogous to the cold-start SFT phase in the DeepSeek-R1 training pipeline. Under this view, applying RL refinement to a distilled reasoning model further enhances its reasoning capabilities, mirroring the multi-stage training dynamics observed in the R1 approach. Enhancing a small, distilled LRM's reasoning performance through RLFT was also conducted in the STILL-3 project \cite{chen_empirical_2025}.

Another effort toward efficient LRM development, in tandem with RLFT of denser base models, has been to utilize PEFT during post-training. Most prominently, Thinking Machines \cite{schulman_lora_2025} stated that Low Rank Adaption (LoRA) \cite{hu_lora_2021} performs equivalently to full fine-tuning for reinforcement learning, and especially highlights how applying LoRA layers over the frozen base model weights provides sufficient tunable capacity to absorb all information learned during RLVR. Combining these ideas of computationally efficient reasoning-model development, the Tina project \cite{wang_tina_2025} unifies both lines of progress by applying parameter-efficient reinforcement learning (PERL) \cite{sidahmed_parameter_2024} directly to a small, distilled LRM. Building on the assumption that the post-training of the distilled LRM is analogous to the cold-start SFT stage in the R1 pipeline, Tina adopts a compact reasoning model as a frozen backbone. It then leverages the insight that LoRA layers provide sufficient tunable capacity for RLVR, enabling RL to shape the model’s reasoning behavior by updating the applied LoRA layers as the model's only trainable parameters. The results from this project demonstrate that the reasoning capabilities of dense LRMs on mathematical problem-solving tasks can be enhanced through a lightweight, parameter-efficient post-training pipeline.

Collectively, these results demonstrate that robust reasoning can be enhanced in compact, distilled LRMs when equipped with an efficient post-training pipeline. Motivated by this shift, we modify the central question explored in this study to the following: Can small, distilled LRMs, when trained using parameter-efficient reinforcement learning with verifiable rewards fine-tuning (PE-RLVR-FT), achieve meaningful improvements on standardized mechanics problems?

\subsection{Outline of Paper}
\label{subsec:paper_outline}

The remainder of this paper is structured as follows. We start of Section~\ref{sec:results} by describing the construction of a synthetic beam mechanics dataset with verifiable ground-truth solutions in Section~\ref{subsec:Dataset_method}. Then, Section~\ref{subsec:RL_method} presents the PE-RLVR-FT pipeline used to adapt a compact, distilled LRM to the task of computing beam support reactions, together with the training results. Subsequently, in Section~\ref{subsec:Eval_method}, we evaluate the resulting BeamPERL model on held-out in-distribution and selected out-of-distribution beam configurations to assess task performance and generalization. We further analyze how task-specific fine-tuning affects broader reasoning ability by evaluating selected training checkpoints on standard mathematical reasoning benchmarks. In Section~\ref{subsec:Discussion}, we discuss the observed training regimes, specialization–robustness trade-offs, and implications for efficient engineering reasoning models. Finally, Section~\ref{sec:conclusion} concludes the paper and outlines potential avenues for future work, while Section~\ref{sec:methods} provides additional details on the experimental methods and materials used in this study.

\section{Results and Discussion}
\label{sec:results}
This study consists of three main components. First, we develop a synthetic dataset of beam mechanics specific question-answer pairs. Second, we PE-RLVR-FT a dense, distilled LRM using the constructed dataset. Lastly, we assess performance on held-out in-distribution data drawn from the same generative process as the training set, as well as out-of-distribution examples designed to test generalization beyond the training support. We also evaluate the model on standard mathematical reasoning benchmarks to analyze how task specialization interacts with general reasoning ability, and to quantify potential catastrophic forgetting.

\subsection{Problem Definition and Dataset Generation}
\label{subsec:Dataset_method}

We focus on beam mechanics problem-solving, a domain grounded in structured physical principles, symbolic manipulation, and verifiable mathematical relationships, and we explore how dense, distilled LRMs can serve as the starting point of self-taught support reaction calculators. Hence, we develop a dataset of synthetic question-answer pairs related to calculating the reaction forces of a statically loaded, simply supported beam.

\begin{figure}[!t]
  \centering
  \includegraphics[width=0.5\textwidth]{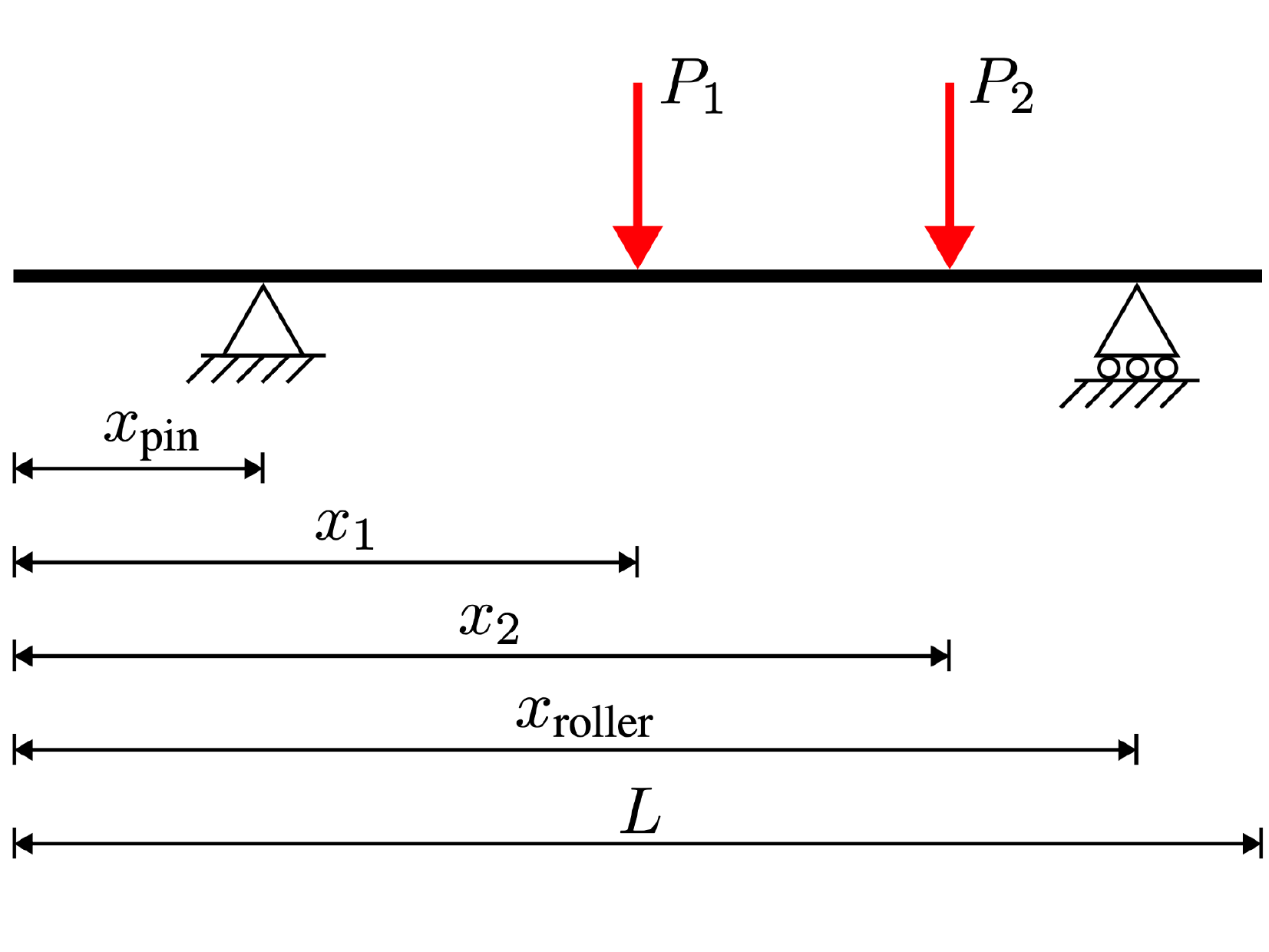}
  \caption{Schematic of a simply supported beam defined on the one-dimensional domain $x\in [0, L]$, where $x$ denotes the axial coordinate along the beam and $L$ is the beam length. The beam is supported by a pinned support located at $x_{\text{pin}} \in [0, L]$ and a roller support located at $x_{\text{roller}} \in [0, L]$, with $x_{\text{pin}} \neq x_{\text{roller}}$. The beam is subjected to $N$ transverse point loads $P_i$, applied at positions $x_i \in [0, L]$, with $x_i \neq x_j \; \forall \; i \neq j$, for $i = 1, \dots, N$, where in this schematic $N=2$.}
  \label{fig:beam}
\end{figure}

We define the beam as a one-dimensional continuum on the span $x\in[0,L]$, where $x$ is a point on the beam and $L$ is the length of the beam. The beam have material properties Young's modulus $E$ and area moment of inertia $I$. The beam is simply supported by a pinned support located at $x_\text{pin}\in[0,L]$ and a roller support located at $x_\text{roller}\in[0,L]$, where $x_\text{pin}\neq x_\text{roller}$. The beam is loaded with $N$ point loads $P_{i}$ located at $x_i\in[0,L],\,$ with $x_i \neq x_j \; \forall \; i \neq j$ and $i=1,2,\dots,N$, each applied normally on the beam. As such, any beam is defined by the set of parameters $\{L,E,I,x_\text{pin},x_\text{roller},\textbf{x},\textbf{P}\}$. A schematic of an arbitrary beam is shown in Figure~\ref{fig:beam}.

We confine ourselves to the first step of beam statics, which is calculating the support reactions of the simply supported beam. The pinned support restricts horizontal and vertical displacements at the point of the support and thus provides the beam with one horizontal and one vertical support reaction, denoted by $R_{H,\text{pin}}$ and $R_{V,\text{pin}}$, respectively. The roller support restricts vertical displacements, i.e., deflections, at the point of the support and thus provides the beam with one vertical support reaction, denoted by $R_{V,\text{roller}}$. Calculating the three support reactions $R_{H,\text{pin}},R_{V,\text{pin}}$ and $R_{V,\text{roller}}$ is a multi-step process, which consists of fulfilling the problem's three equilibrium equations, the sum of forces in the horizontal direction, the sum of forces in the vertical direction and the sum of moments along the beam, all which needs to equal zero to satisfy the static behavior of the beam. The three equations are defined in Equation~\ref{eq:equi_equa}, respectively.

\begin{equation}
\label{eq:equi_equa}
\sum F_H = 0\;, \quad
\sum F_V = 0\;, \quad
\sum M \ = 0
\end{equation}

Since no loads are applied in the horizontal direction, the horizontal equilibrium condition in Equation~\ref{eq:equi_equa} is trivially satisfied, yielding \(R_{H,\text{pin}} = 0\). The vertical force equilibrium equation and the moment equilibrium equation are then enforced by substituting the $N$ applied loads and the two unknown vertical support reactions, and solving the resulting system of equations for the two unknown vertical reaction forces, $R_{V,\text{pin}}$ and $R_{V,\text{roller}}$, stated below.

\begin{subequations}\label{eq:equi_equa_inserted}
\begin{align}
R_{V,\text{pin}} + R_{V,\text{roller}} + &\sum_{i=1}^{N} P_i = 0 \label{eq:equi_equa_inserted:ver_equi} \\
R_{V,\text{pin}}(x_\text{pin}-x_\text{j}) + R_{V,\text{roller}}(x_\text{roller}-x_\text{j}) + &\sum_{i=1}^{N} P_i(x_\text{i}-x_\text{j}) = 0 \quad \forall \ x_j\in[0,L] \label{eq:equi_equa_inserted:mom_equi}
\end{align}
\end{subequations}

For the training data we define a subset of the beams as those with lengths $L\in\{\,l,\,2l,\,3l\,\}$, Young's modulus $E$, area moment of inertia $I$, support locations $x_\text{pin}=0$ and $x_\text{roller}=L$, and loaded with a single point load $P\in\{-p,-2p,-3p\}$ at a location $x_P\in\{\, 0.05\,k\,L \mid k = 0,1,2,\dots,20 \,\}$. Here $l,E,I,p$ are symbolic, unitless magnitudes and the minus sign in front of the point load magnitude signifies that the load is applied on the beam normally in the downward direction. As such, the training data consists of 189 distinct statically loaded beam configurations.

\begin{figure}[!t]
  \centering
  \includegraphics[width=0.9\textwidth]{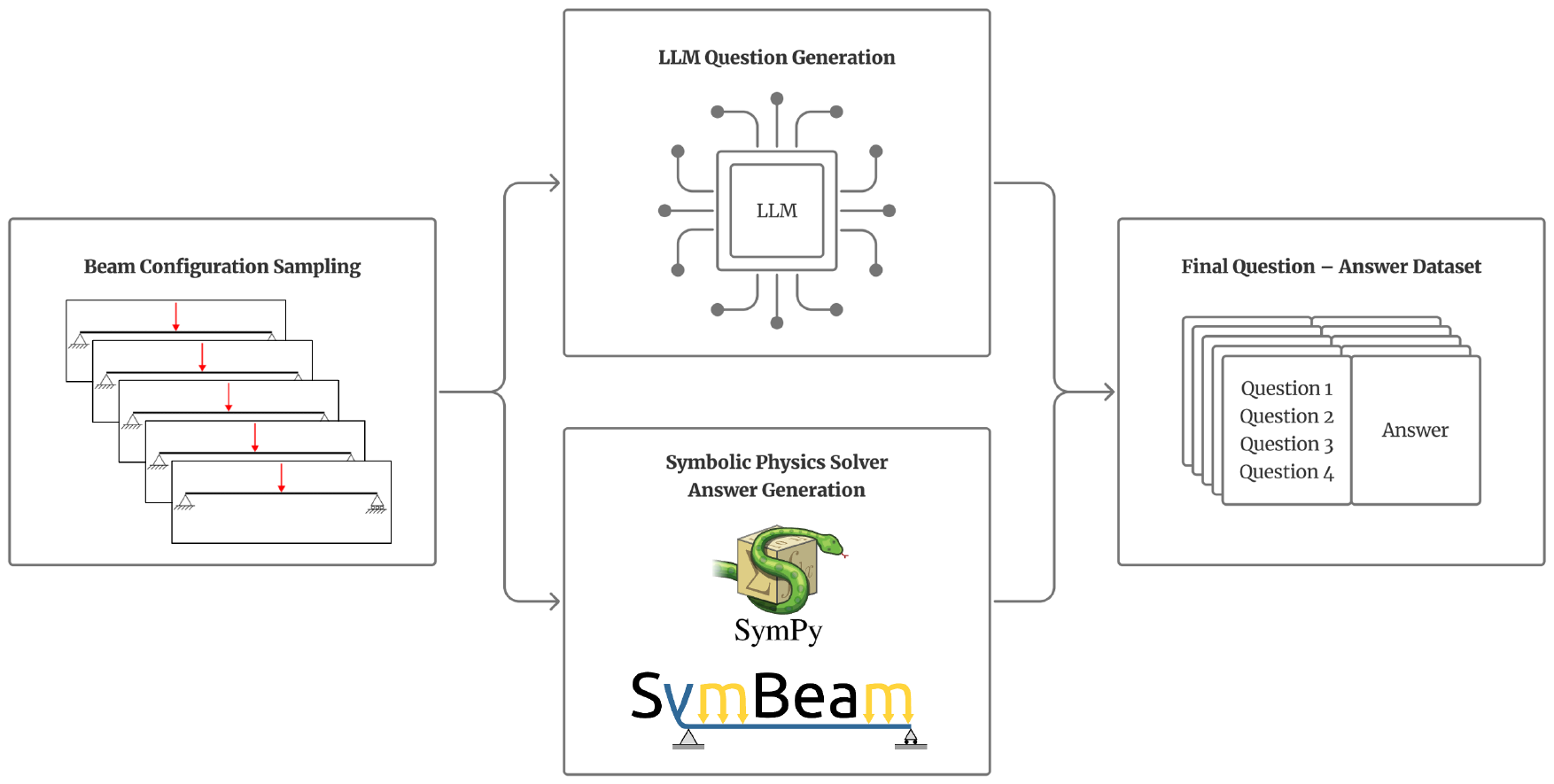}
  \caption{End-to-end dataset generation pipeline for beam-mechanics problem-solving. Discrete beam configurations are first sampled from a symbolic parameter space and solved analytically using a symbolic mechanics solver to obtain exact ground-truth support reactions. An LLM is used to generate multiple natural-language problem formulations for each beam configuration, while the underlying physical solution remains unchanged. This many-to-one construction yields a verifiable question–answer dataset with multiple linguistic variants mapped to a single mechanically correct solution.}
  \label{fig:data_gen}
\end{figure}

We use a modified version of the Python library \texttt{SymBeam} \cite{carneiro_symbeam_2020}, which leverages the symbolic engine \texttt{SymPy} \cite{meurer_sympy_2017}, to solve the static beam problem. For each beam configuration, the solver produces analytical expressions and corresponding numerical evaluations of the external reaction forces, internal force and moment distributions, as well as the deflection and rotation fields along the beam's axis. We use an LLM to generate self-contained questions that fully specify each beam configuration and ask the reader to compute the corresponding support reactions. For each beam configuration, four distinct question formulations are generated that map to the same ground-truth solution, resulting in a total of 756 question–answer pairs in the training dataset. Figure~\ref{fig:data_gen} shows a high level overview of the end-to-end data generation pipeline used to produce verifiable question–answer pairs for beam-mechanics problem-solving, while a representative example from the training dataset is shown in the Text Box~\ref{textbox:Example_QAPair}. More detailed information on the data generation pipeline is provided in Section~\ref{subsec:dataset}.

\begin{tcolorbox}[
  title={Example Question--Answer Pair},
  label={textbox:QA_example},
  colback=gray!5,
  colframe=black,
  fonttitle=\bfseries
]
\textbf{Question:} 

Given a beam of length $2*L$ with a pin support at $x=0$ and a roller support at $x=2*L$, and a downward point load of $-3*P$ at $x=0.9*L$, calculate the reaction forces at the supports. The beam has a Young's modulus of $E$ and a moment of inertia of $I$.

\vspace{0.5em}
\textbf{Answer:}

$\left[\,1.65P,\,1.35P\,\right]$
\end{tcolorbox}
\noindent\begin{minipage}{\textwidth}
\captionof{boxes}{Representative question–answer pair illustrating the format and content of the training data.}
\label{textbox:Example_QAPair}
\end{minipage}

\subsection{Training Strategy}
\label{subsec:RL_method}

Having constructed a dataset of beam-mechanics questions with verifiable ground-truth answers, we fine-tune a dense, distilled LRM using RLVR. The objective of this stage is to adapt a small, general-purpose reasoning model into a task-specific engineering LRM capable of reliably solving beam equilibrium problems. Fine-tuning is performed using GRPO, a policy-gradient–based method that enables reinforcement learning without an explicit value function \cite{shao_deepseekmath_2024}. Instead, GRPO operates by comparing the relative performance of groups of sampled responses, which are ranked according to a predefined reward function. The policy is then optimized by increasing the probability of sampling the higher-ranked responses relative to lower-ranked ones. This makes GRPO well suited for tasks with deterministic, externally verifiable rewards, such as those arising from symbolic mechanics solvers. As a result, effective learning can be realized without the explicit generation or supervision of reasoning traces, which are required in SFT pipelines. Whereas SFT constrains the model to imitate fixed reasoning traces, RLVR permits the model to discover its own internal reasoning strategies, driven by feedback on solution correctness provided by a predetermined reward function.

As highlighted in the development of GRPO \cite{shao_deepseekmath_2024} and further supported by recent findings \cite{yue_does_2025}, RLVR methods that employ binary reward functions amplify reasoning capabilities present in the base model by refining the output distribution, thereby increasing the likelihood that generated token sequences converge to the correct final answer. Consequently, the choice of base model is a critical design decision, as it determines the initial reasoning capacity available to the RL process. This behavior is expected, since the training strategy relies on the model receiving an initial reward signal to guide optimization. A model with no prior knowledge of beam statics is therefore unlikely to generate token sequences that yield correct solutions and may consequently fail to receive informative rewards to kick off the learning process. To address this challenge, we explored several strategies to ensure that the base model used for RLFT possessed sufficient beam mechanics knowledge to successfully initiate learning. Throughout this study, we restrict our attention to dense models with at most 3 billion parameters, consistent with the objective of developing self-taught beam-mechanics reasoning models under restrictive computational constraints. An initial approach was to investigate pure RL applied to dense, pretrained base LLMs. However, due to the limited initial performance of these smaller base LLMs, training struggled to make progress under the sparse reward structure employed.

A second approach involved introducing an initial stage of SFT on synthetically generated and verified question–reasoning trace pairs, analogous to the structured thought integration phase in PRefLexOR~\cite{buehler_preflexor_2025}. While effective in theory, this approach requires distilling reasoning traces from larger LRMs and performing an additional supervised training stage. This increased the complexity and computational cost of the training pipeline, but more importantly, as PRefLexOR itself demonstrated, anchoring the model to explicit reasoning trajectories during an initial phase is most effective when followed by a second phase that masks those traces and forces the model to develop independent reasoning, suggesting that the reasoning traces serve primarily as scaffolding rather than as the final learning objective. This insight, combined with the availability of distilled LRMs that have already undergone extensive reasoning-oriented pretraining, motivated our decision to bypass the SFT stage entirely and proceed directly with RLVR applied to a distilled base model. 
Inspired by recent work \cite{dang_reinforcement_2025, wang_tina_2025,chen_empirical_2025}, we adapted our training strategy to better leverage recent advancements in open-source, distilled LRMs. Rather than introducing an initial stage of SFT on synthetic, beam-mechanics-specific reasoning traces, we initialize training from a distilled LRM that has already been pretrained on a large corpus of curated reasoning prompts and verified reasoning trajectories spanning a broad range of domains, including mathematical reasoning. We adopt this approach under the assumption that such a model possesses sufficient prior knowledge to receive informative initial reward signals and initiate the learning process without requiring an additional round of extensive domain-specific SFT. As a result, throughout post-training the self-taught beam mechanics reasoning model is never exposed to explicit reasoning traces demonstrating how to solve the posed questions. Instead, learning is driven solely by binary correctness rewards on the final answer, supplemented by a lightweight format reward. This training setup allows the model to freely explore its own solution strategies, rather than being constrained to predefined reasoning paths.

\begin{figure}[!b]
  \centering
  \includegraphics[width=0.99\textwidth]{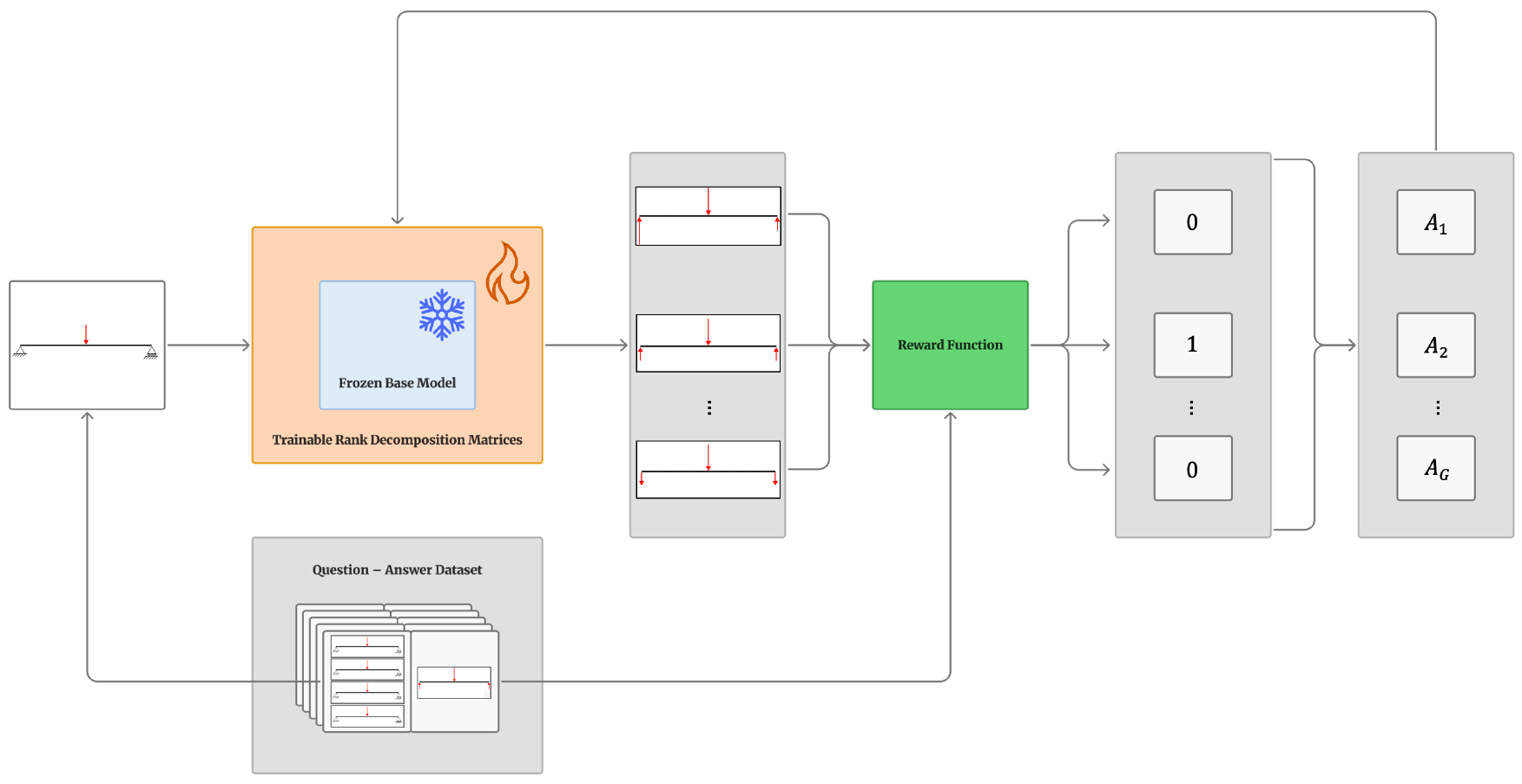}
  \caption{PE-RLVR-FT workflow for adapting a distilled LRM to beam-mechanics problem-solving. Beam-mechanics questions from the synthetic dataset are used to prompt a frozen base model augmented with trainable LoRA adapters. For each prompt, the model samples a group of $G$ candidate responses, which are evaluated by a deterministic reward function based on format adherence and symbolic beam statics correctness, each assigning a binary reward. These rewards are converted into relative advantage signals $A_1,\dots,A_G$ and used by GRPO to update only the LoRA parameters, while all pretrained backbone weights remain frozen. A more detailed description of this process is provided in Section~\ref{subsec:methods_training}. Adapted from \cite{deepseek-ai_deepseek-r1_2025}.}
  \label{fig:training_flowchart}
\end{figure}

To enable computationally efficient post-training, we employ PEFT by restricting parameter updates to applied LoRA adapters inserted into the base model, while keeping all pretrained backbone weights frozen. The training pipeline builds upon the Tina project \cite{wang_tina_2025}, which extends the Open R1 framework \cite{hugging_face_open_2025} to support PERL with GRPO-based optimization and LoRA-based adaptation. Figure~\ref{fig:training_flowchart} summarizes the model development and training workflow used to adapt a distilled LRM into a task-specific engineering reasoning model. Additional implementation and training details are provided in Section~\ref{subsec:methods_training}.

\subsubsection{Training Results}

We initialize training from the \texttt{DeepSeek-R1-Distill-Qwen-1.5B} model, the smallest distilled LRM in the DeepSeek-R1 model family \cite{deepseek-ai_deepseek-r1_2025}. During training, we monitor metrics including training rewards, approximated Kullback–Leibler (KL) divergence, and completion length to better understand how fine-tuning affects model behavior. Checkpoints of the fine-tuned model are saved at multiple stages throughout training for subsequent evaluation. Additional training details are provided in Section~\ref{subsec:methods_training}.

\begin{figure}[!b]
  \centering
  \begin{subfigure}[b]{0.49\textwidth}
    \centering
    \includegraphics[width=\textwidth]{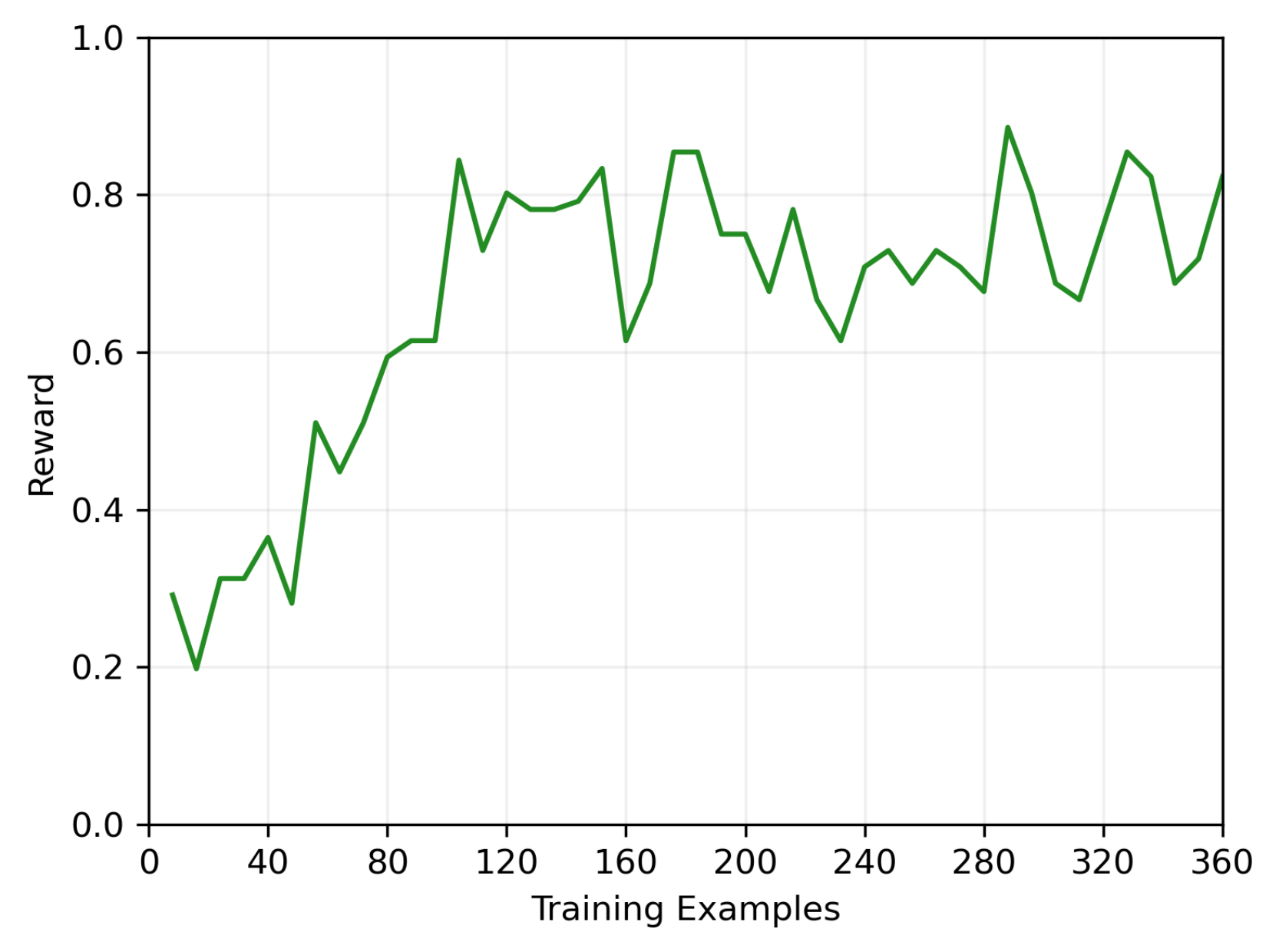}
    \caption{}
    \label{fig:reward}
  \end{subfigure}
  \hfill
  \begin{subfigure}[b]{0.49\textwidth}
    \centering
    \includegraphics[width=\textwidth]{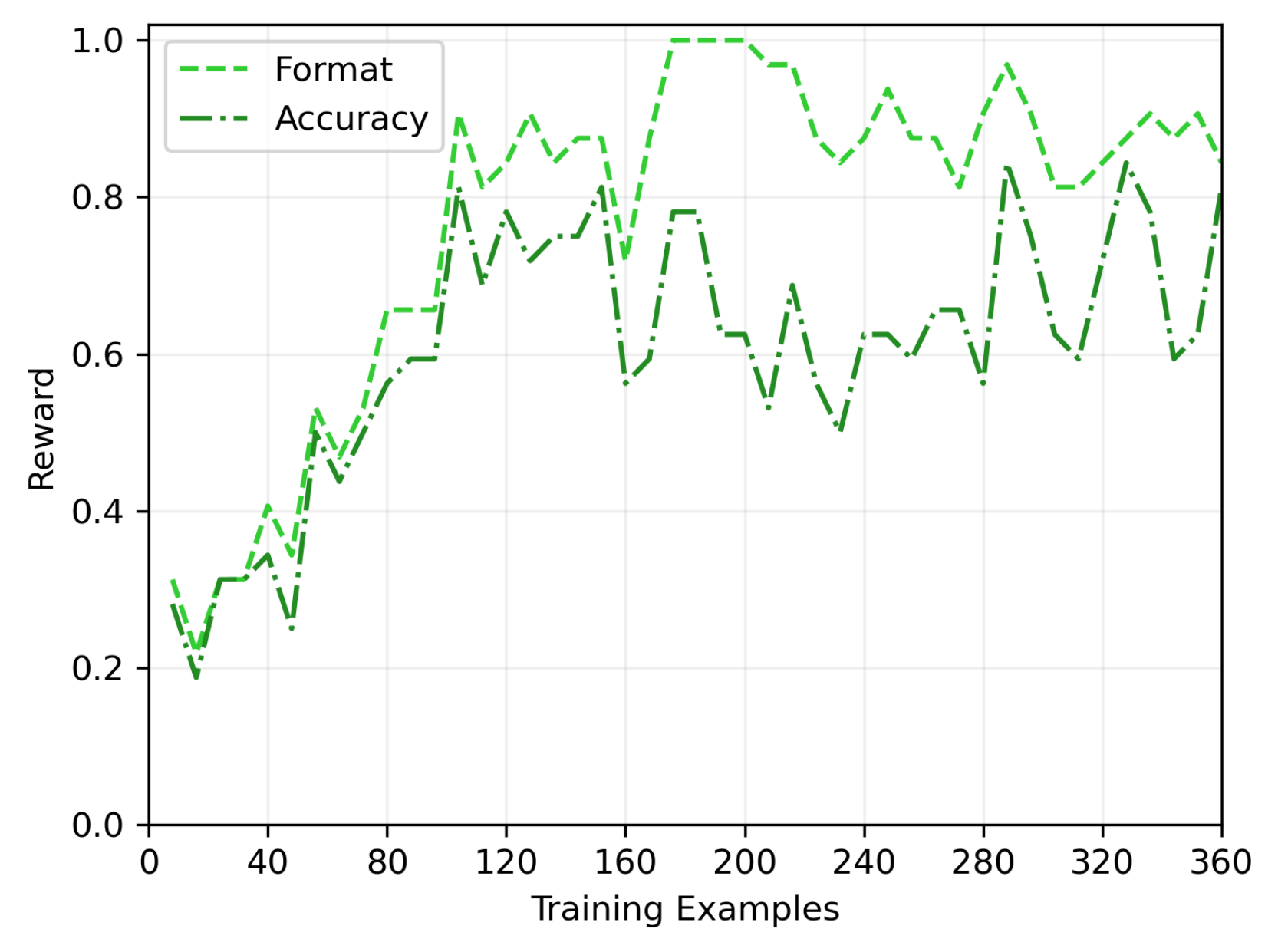}
    \caption{}
    \label{fig:format_n_accuracy}
  \end{subfigure}
  \caption{Training performance against the cumulative number of training examples. (a) Total weighted reward. (b) Unscaled format and accuracy rewards. All rewards are averaged across all rollouts for all observed examples at each training step. Both sub-rewards increase sharply during the early training phase, after which the accuracy reward slightly decreases while the format reward remains consistently high, indicating that early performance gains are primarily associated with improved output formatting.}
  \label{fig:reward_format_n_accuracy}
\end{figure}

Figure~\ref{fig:reward} shows the average reward across all rollouts and training examples at each training step, plotted against the cumulative number of training examples. We observe a sharp increase in training reward from approximately 0.2 to 0.8 over the first 120 training examples, where 0 and 1 are the lowest and highest possible scores, respectively, after which the reward plateaus and remains approximately constant for the remainder of the reinforcement learning stage, with intermittent spikes between roughly 0.6 and 0.85. Figure~\ref{fig:format_n_accuracy} decomposes the total reward into format and accuracy components, revealing that both sub-rewards increase concurrently during the initial training phase. After approximately 120 training examples, the accuracy reward decreases slightly, albeit with large variance, while the format reward remains consistently above 0.8.

\begin{figure}[!t]
  \centering
  \begin{subfigure}[b]{0.49\textwidth}
    \centering
    \includegraphics[width=\textwidth]{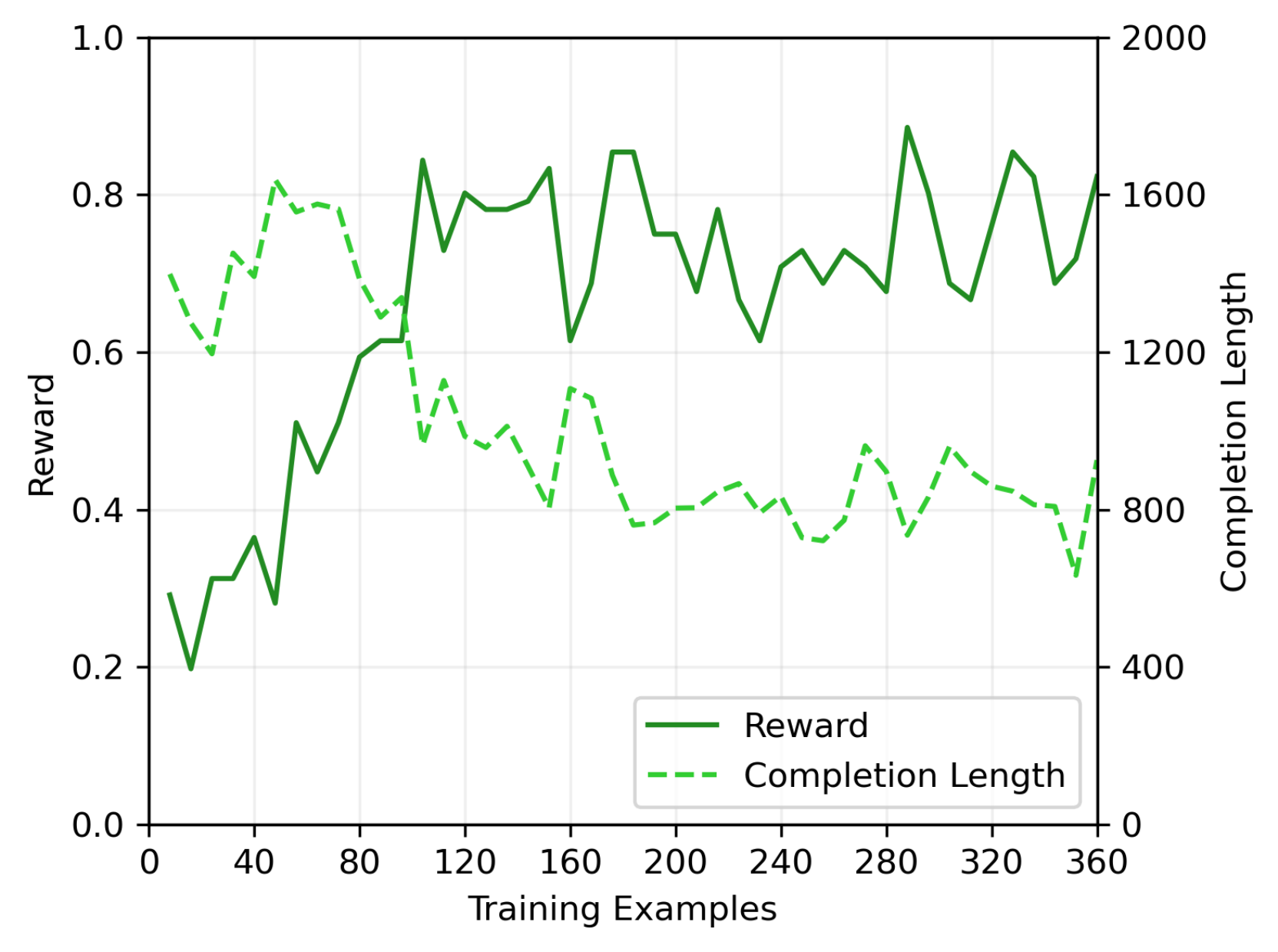}
    \caption{}
    \label{fig:reward_n_completionlength}
  \end{subfigure}
  \hfill
  \begin{subfigure}[b]{0.49\textwidth}
    \centering
    \includegraphics[width=\textwidth]{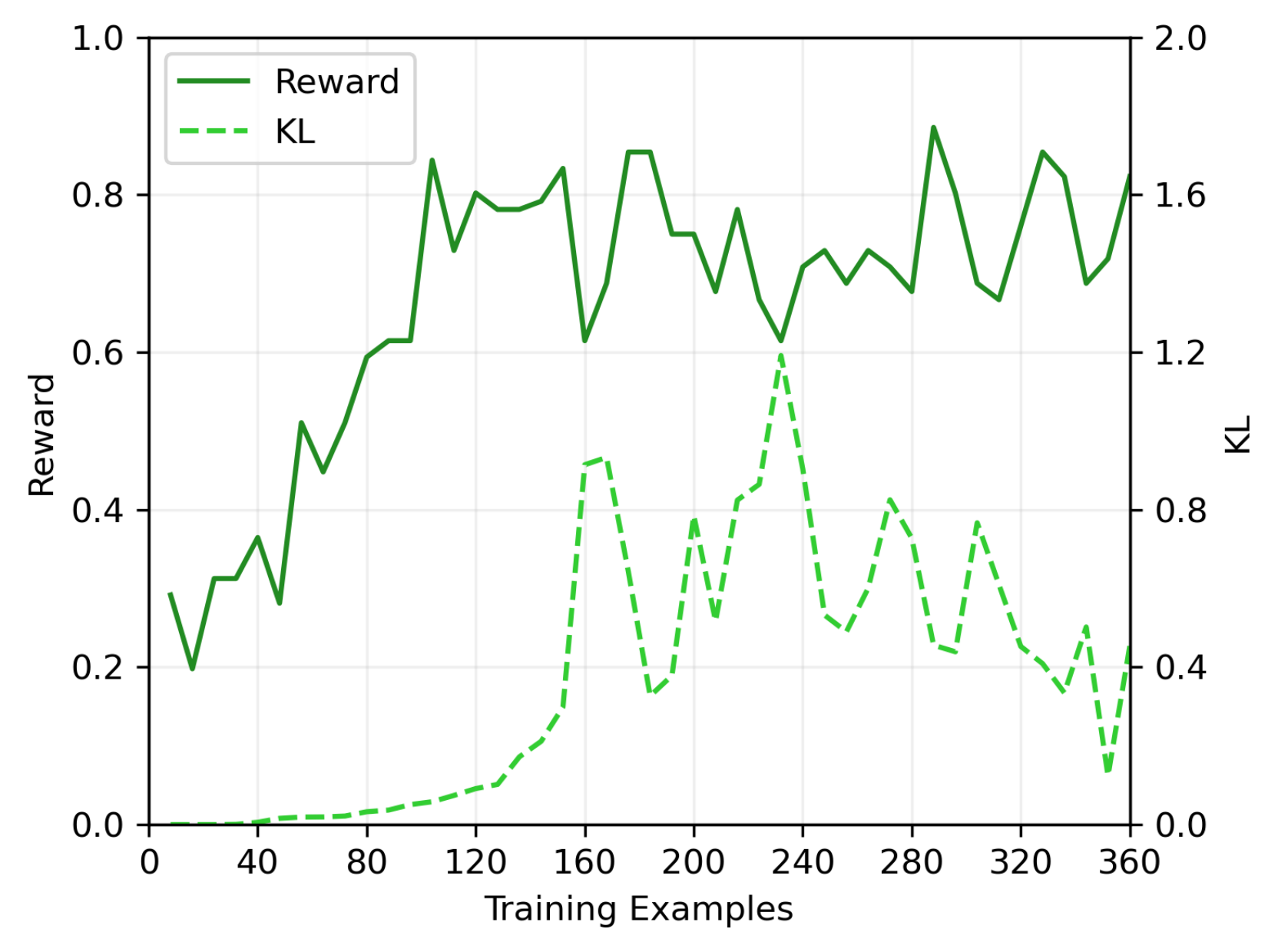}
    \caption{}
    \label{fig:reward_n_kl}
  \end{subfigure}
  \caption{Evolution of training reward, completion length, and KL divergence plotted against the cumulative number of training examples. In (a), reward and rollout completion length are shown on separate y-axes, with completion length, in tokens, decreasing early in training and stabilizing at a task-appropriate token length. In (b), reward and KL divergence are shown on separate y-axes, with KL divergence increasing conservatively at first and more sharply at later stages. All values are averaged across all rollouts for all observed examples at each training step.}
  \label{fig:reward_vs_completion_n_kl}
\end{figure}

These results suggest that early performance gains are primarily driven by improved output formatting, with the model learning to structure its responses in a way that enables successful problem solving. Figure~\ref{fig:reward_n_completionlength} further illustrates this behavior by plotting both the average reward and the average completion length per rollout against the cumulative number of training examples. We observe that completion length decreases during the initial training phase, reaching a task-appropriate token length, and then remains stable for the remainder of training. Notably, the increase in reward is achieved without an increase in token usage, indicating that the model learns to solve the task more efficiently.

A central question that arises is why training plateaus after approximately 120 training examples, despite the diversity of the training data and the absence of repeated examples. We hypothesize that this behavior reflects a fundamental transition in the model.  During the initial phase, the model primarily refines the baseline LRM’s response patterns to better align with the task requirements. Beyond this point, we observe a sharp increase in KL divergence, suggesting that the model begins to deviate more substantially from the baseline LRM, resulting in diminished marginal gains from further training. This behavior is illustrated in Figure~\ref{fig:reward_vs_completion_n_kl}, which plots both the average reward and the average KL divergence per rollout against the cumulative number of training examples. To further investigate this hypothesis, we evaluate models saved at evenly spaced checkpoints throughout training. This evaluation examines whether and how the model’s problem-solving strategy evolves over the course of training, including its behavior on tasks that differ from those encountered during training. The results of this analysis are presented in the following section.

\subsection{Evaluation Strategy}
\label{subsec:Eval_method}

To evaluate the fine-tuned model, we construct an evaluation dataset comprising held-out in-distribution (ID) and out-of-distribution (OOD) beam mechanics examples. The model is evaluated at 10 checkpoints, evenly spaced throughout training, from the initialized base model to the final fine-tuned model. At each checkpoint, we evaluate performance on 24 evaluation samples, generating seven outputs per sample. For each output, we assess format adherence and accuracy, and compute pass@1, pass@7 and majority@7 accuracy metrics. In addition, selected checkpoints are evaluated on standard mathematical reasoning benchmarks to assess how task specialization affects general reasoning ability and to quantify potential catastrophic forgetting. Further information on the evaluation strategy, such as precise definitions of accuracy, output parsing, and decoding and sampling settings are provided in Section~\ref{subsec:evaluation}.

\subsubsection{Evaluation Data}
\label{subsec:Eval_data}

We construct the custom evaluation dataset using the same generative process as the training set. While any natural-language question not appearing verbatim in the training data could be considered held out, we adopt a stricter definition based on the underlying physical parameters of each beam: an evaluation example is considered held out only if its combination of beam parameters does not occur in the training data. We impose an even stronger restriction to explicitly evaluate the model’s ability to generalize its reasoning beyond memorized parameter configurations. For all beam examples in the evaluation set, the beam length satisfies $L\notin\{\,l,\,2l,\,3l\,\}$, the point-loads satisfy $P_i\notin\{-p,-2p,-3p\}$ and the point-load locations satisfy $x_i\notin\{\, 0.05\,k\,L \mid k = 0,1,2,\dots,20 \,\}$. The Young’s modulus and area moment of inertia are kept fixed across training and evaluation datasets.

ID evaluation data consist of beams with support locations located at the beam ends, $(\,x_\text{pin},\,x_\text{roller}\,)=(\,0,\,L\,)$, and subjected to a single point load $P$. OOD data include beams that violate either of these conditions. Specifically, OOD examples either (i) contain multiple point loads $P_i, \; i \in \{1, \dots, N\}$, or (ii) have support locations that are not both at the beam ends, that is, $(\,x_{\text{pin}},\, x_{\text{roller}}\,) \in \{(a, L), (0, b), (a, b) \mid (a, b) \neq (0, L)\}$. We limit the evaluation data to beams with parameters $(\,L,\,P_i\,)=(\,9l,\,-13p\,)$. For beams with supports located at the beam ends, we consider configurations with $N=1,2,3$ applied point loads. Each point load is placed at a unique location chosen from $x_i\in\{\,\frac{1}{8}L,\,\frac{1}{3}L,\frac{21}{40}L,\,\frac{2}{3}L\,\}$. For beams in which one or both supports are not located at the beam ends, we consider configurations with possible support locations $(\,x_{\text{pin}},\, x_{\text{roller}}\,) \in \{(0.1L, L), (0, 0.9L), (0.1L, 0.9L)\}$ and $N=1,2$ applied point loads. In these cases, each point load is placed at a unique location chosen from either free ends of the beam or the midpoint between the two supports (i.e., the midspan of the supported segment, not the midspan of the full beam).

The resulting evaluation dataset comprises 24 samples, and while the evaluation set is small in absolute size, it is deliberately constructed to span distinct structural regimes, enabling controlled analysis of generalization rather than statistical performance estimation. These are subdivided into 4 held-out ID examples of simply supported beams with a single applied point load, 8 OOD examples of simply supported beams with multiple applied point loads, and 12 OOD examples of beams with one or two cantilevered ends, subjected to one or two point loads located either at the midspan of the supported segment or at the cantilever tip. Additional details of the evaluation dataset are provided in Section~\ref{subsec:dataset}.

\subsubsection{Evaluation Results}

\begin{table}[!b]
\centering
\caption{Performance comparison between the base model and BeamPERL (best-performing checkpoint) across the evaluation metrics. Results are reported in terms of Pass@1, Pass@7, and Majority@7 accuracy, allowing for one-shot, best-case, and output consistency performance analysis.}
\label{tab:base_vs_best}
\begin{tabular}{l|ccc}
\toprule
\textbf{Model} & \textbf{Pass@1} & \textbf{Pass@7} & \textbf{Maj@7} \\
\midrule
\texttt{DeepSeek-R1-Distill-Qwen-1.5B} & $12.50\%$ & $29.17\%$ & $0.00\%$ \\
\midrule
\texttt{BeamPERL} (best-performing checkpoint) & $20.83\%$ & $41.67\%$ & $4.17\%$ \\
\bottomrule
\end{tabular}
\end{table}

\begin{figure}[!b]
  \centering
  \includegraphics[width=0.5\textwidth]{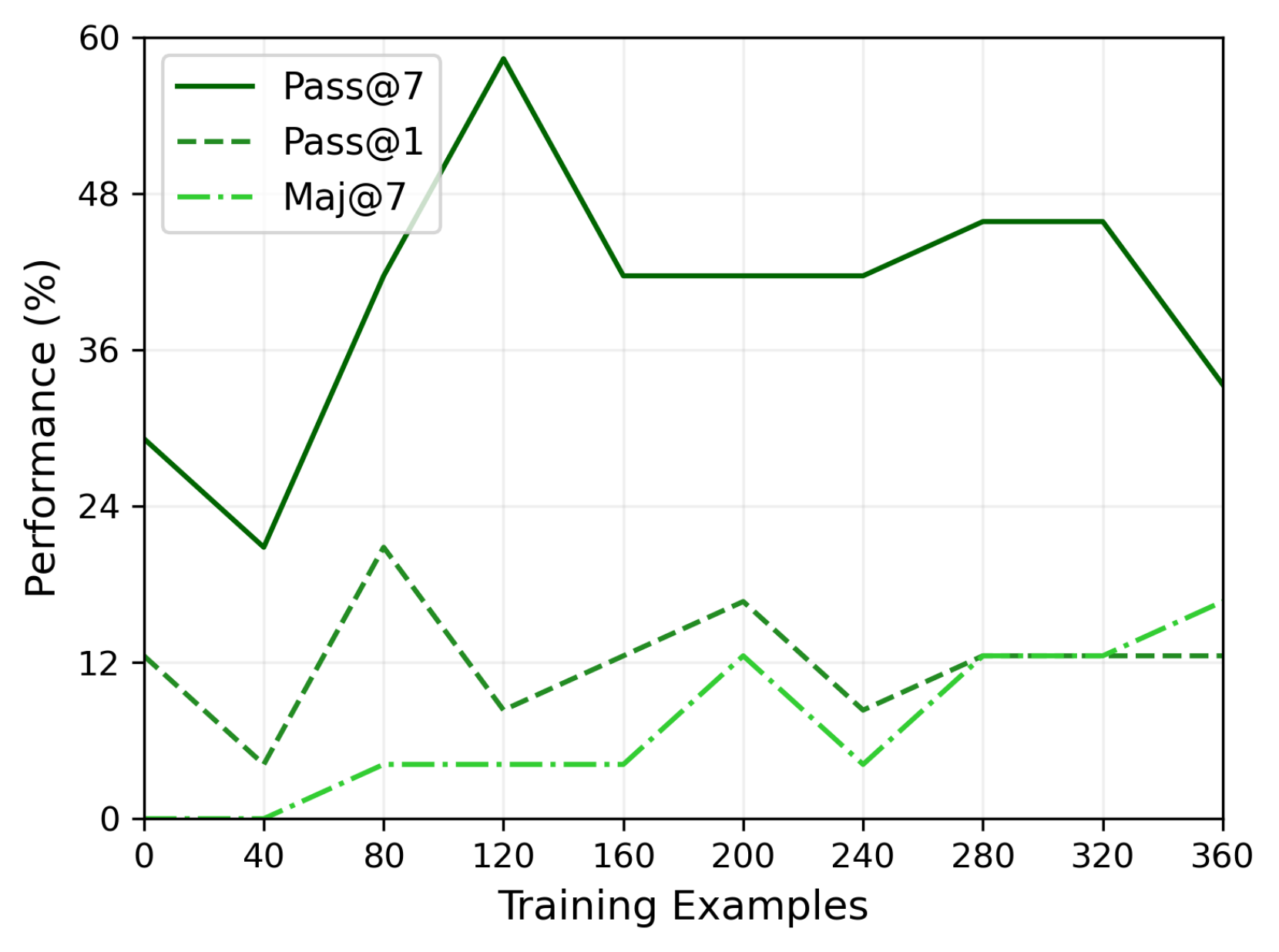}
  \caption{Evaluation performance across ten equally spaced training checkpoints. Each metric is averaged across all ID and OOD examples. Pass@1 and Pass@7 show that both one-shot and best-case performance initially increase before a slight subsequent decline, while Majority@7 shows a steady, gradual improvement throughout training.}
  \label{fig:pass1_pass7_maj7}
\end{figure}

We now present the evaluation results of the PE-RLVR-FT BeamPERL model across training. Results are reported in terms of Pass@1, Pass@7, and Majority@7 accuracy, allowing us to analyze one-shot and best-case performance, as well as output consistency as the model is fine-tuned. By tracking these metrics across evenly spaced checkpoints, we characterize how performance evolves during training and identify the regime in which fine-tuning is most effective. Table~\ref{tab:base_vs_best} shows that the best-performing checkpoint of the PE-RLVR-FT BeamPERL model achieves a $66.7\%$ improvement in Pass@1 and a $42.9\%$ improvement in Pass@7 relative to the base model, while increasing the Majority@7 score from $0\%$ to $4.17\%$. Figure~\ref{fig:pass1_pass7_maj7} illustrates the evolution of Pass@1, Pass@7, and Majority@7 performance across the ten recorded checkpoints during training. Performance improves rapidly in the early stages of fine-tuning, peaks after 80–120 training examples, plateaus, and subsequently declines at later stages of training.

\begin{figure}[!t]
  \centering
  \begin{subfigure}[b]{0.32\textwidth}
    \centering
    \includegraphics[width=\textwidth]{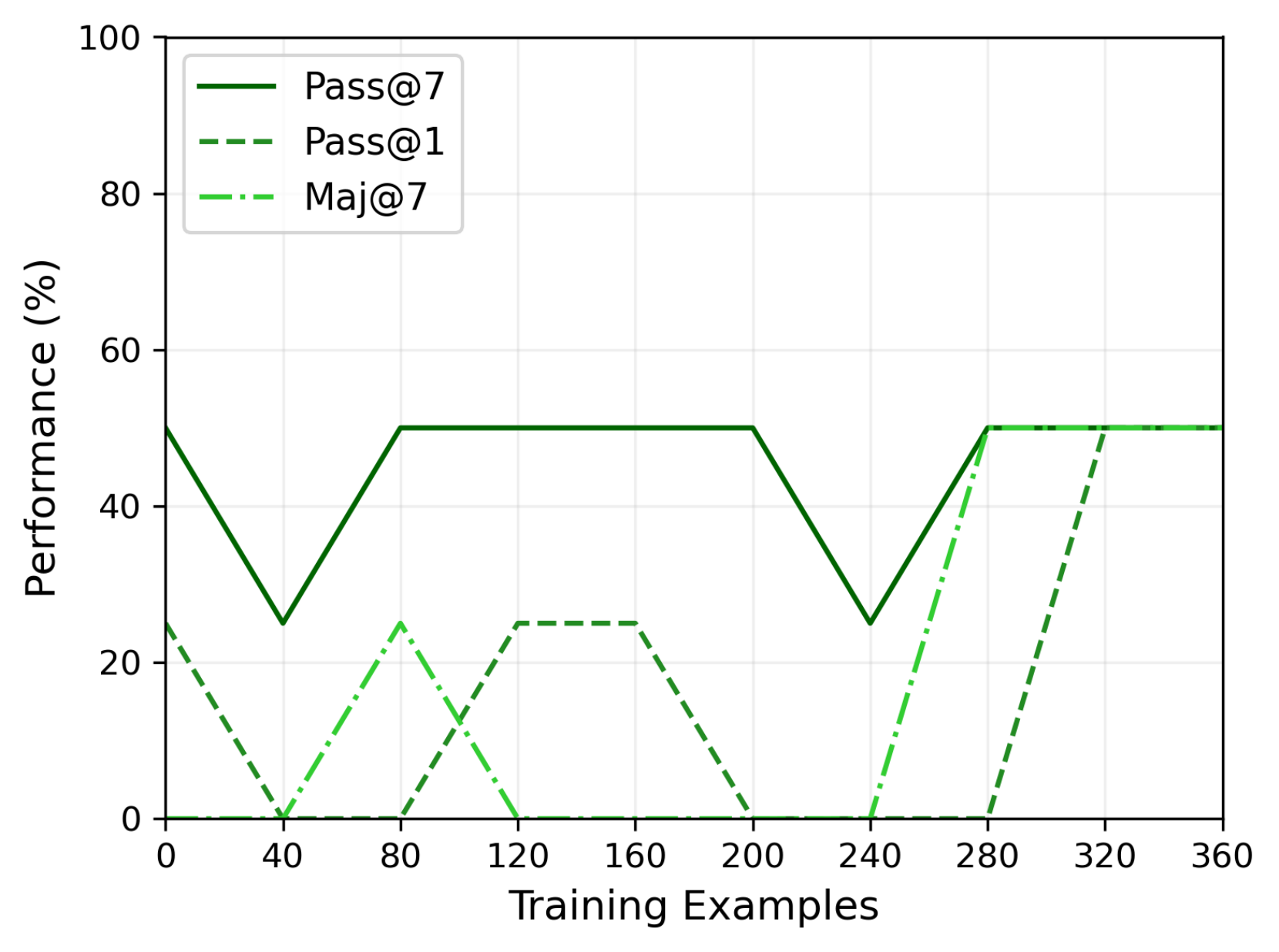}
    \caption{}
    \label{fig:pass_first4}
  \end{subfigure}
  \hfill
  \begin{subfigure}[b]{0.32\textwidth}
    \centering
    \includegraphics[width=\textwidth]{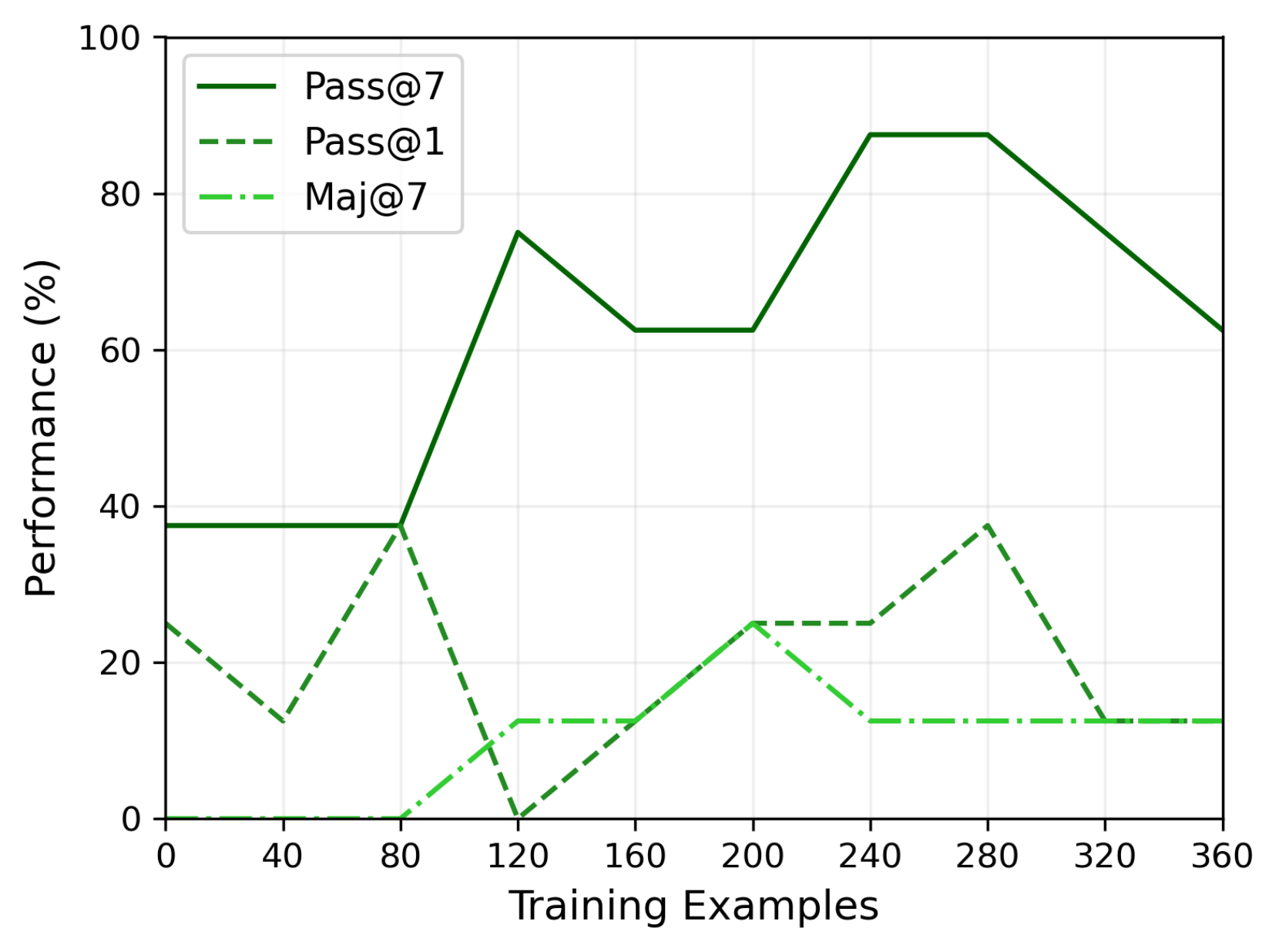}
    \caption{}
    \label{fig:pass_mid8}
  \end{subfigure}
  \hfill
  \begin{subfigure}[b]{0.32\textwidth}
    \centering
    \includegraphics[width=\textwidth]{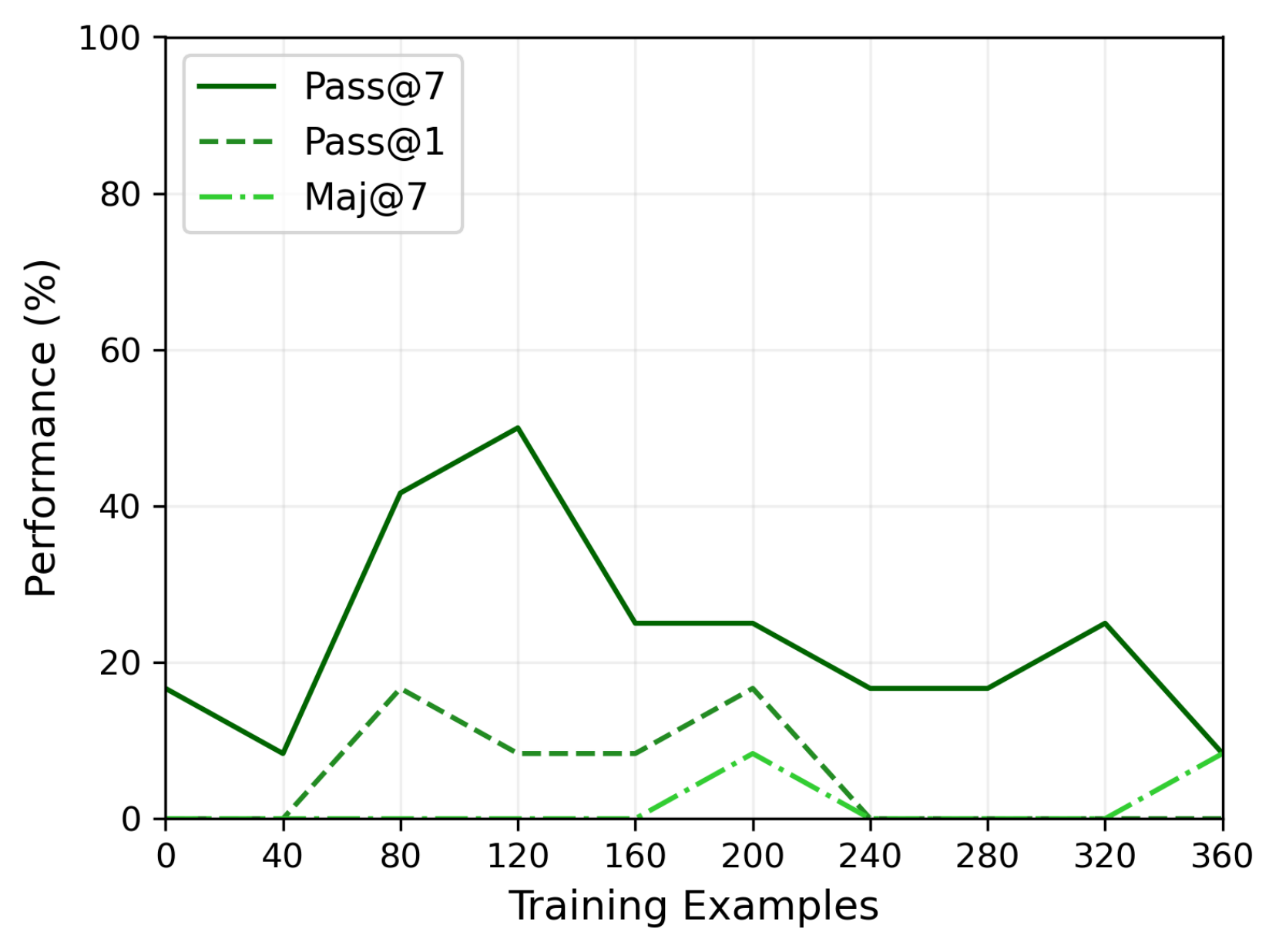}
    \caption{}
    \label{fig:pass_last12}
  \end{subfigure}
  \caption{Evaluation performance partitioned across three plots: (a) ID samples consisting of beams supported at the ends with a single load, (b) OOD samples with multiple loads applied at the ends, and (c) OOD samples with varying support locations. Each metric is averaged over the corresponding evaluation examples. Results show improved one-shot performance and increased output consistency on ID samples as training progresses, while generalization across OOD dimensions differs: performance increases for increasing load multiplicity throughout training but degrades for varying support locations when training extends beyond the optimal regime.}
  \label{fig:pass1_pass7_maj7_comparison}
\end{figure}

By partitioning the evaluation data into ID samples consisting of beams supported at the ends with a single load, OOD samples of beams supported at the ends with multiple loads, and OOD samples with varying support locations, we gain clearer insight into how the model generalizes its beam statics knowledge during training. Figure~\ref{fig:pass1_pass7_maj7_comparison} presents this partitioning across three plots, showing Pass@1, Pass@7, and Majority@7 performance, each averaged over the corresponding evaluation example types. Across the four ID beam samples, Pass@7 remains approximately constant at $50\%$ throughout training, while both Pass@1 and Majority@7 increase and converge to $50\%$ by the end of training. This indicates improved one-shot performance and increased output consistency on ID samples as training progresses. For the two types of OOD samples, we observe contrasting trends. While Pass@7 performance on OOD samples consisting of beams supported at the ends with multiple loads increases steadily throughout training, performance on OOD samples with varying support locations peaks after approximately 80–120 training examples and subsequently declines. This indicates that fine-tuning improves the model’s ability to generalize along certain OOD dimensions, such as the number of applied loads, while potentially inducing overfitting along other dimensions, notably variations in support location, when training extends beyond the optimal regime. These results motivate a more detailed analysis of the model’s generated rollouts beyond aggregate performance metrics. In particular, examining how reasoning traces and final answers evolve throughout training provides deeper insight into the mechanisms underlying the observed performance trends.

\subsubsection{Sample Results}

In the Supplementary Information (SI) we present qualitative results for three representative evaluation samples: (i) an ID example, (ii) an OOD example involving beams supported at the ends with multiple applied loads, and (iii) an OOD example with varying support locations. Selected results are also also presented in this section.

For the ID example we present three representative outputs generated by the base model: a fully correct response receiving the maximum reward, also shown in Text Box~\ref{textbox:QA_ID_1_BaseModel_Correct}, a response with correct formatting but an incorrect solution, and a response with a correct solution but incorrect formatting. These outputs illustrate that the base model possesses sufficient prior knowledge to receive informative reward signals, while still exhibiting both formatting and engineering reasoning errors that can be addressed through fine-tuning. We also show a correct response from the best-performing BeamPERL training checkpoint, also shown in Text Box~\ref{textbox:QA_ID_1_BeamPERL_Correct}, and a correct response from the final BeamPERL training checkpoint, indicating that problem solving capabilities on ID data are retained throughout training.

\begin{tcolorbox}[
  title={ID Evaluation Example -- Correct Output Sampled from Base Model},
  colback=gray!5,
  colframe=black,
  fonttitle=\bfseries,
  fontupper=\small,
  breakable
]
\texttt{<think>}

Okay, so I need to figure out the reaction forces at the pin and roller supports for this beam problem. Let me start by understanding the problem setup. The beam is 9*L long, supported by a pin at x=0 and a roller at x=9. There's a point load of -13P applied at 4.725L from the left end.

First, I remember that reaction forces are the forces that counteract the applied loads and moments to keep the beam in equilibrium. Since it's a pin support at x=0, it can have both vertical and horizontal reactions. The roller at x=9L can only provide a vertical reaction because there's no constraint in the horizontal direction.

I think the first step is to write down the equations of equilibrium. For a beam in static equilibrium, the sum of forces in both x and y directions must be zero, and the sum of moments about any point must also be zero.

Let me denote the reaction forces as R\_A at x=0 (pin) and R\_B at x=9L (roller). The point load is -13P, which I assume is a downward force. Since the beam is being supported at both ends, the reactions will have both vertical components.

So, the sum of the vertical forces should be zero:
\[
R_A + R_B = 13P
\]
That's one equation. Now, taking moments about one of the supports should give me another equation. I'll take moments about the pin at x=0 because it's easier since some distances will be zero.

The moment due to R\_B about x=0 is R\_B * 9L. The moment due to the load at 4.725L is (-13P) * 4.725L. Since the moment about the pin should sum to zero:
\[
R_B * 9L = 13P * 4.725L
\]
Dividing both sides by L:
\[
9R_B = 13 * 4.725P
\]

Calculating 13 * 4.725:

13 * 4 = 52

13 * 0.725 = 9.425

So total is 52 + 9.425 = 61.425

Thus:
\[
9R_B = 61.425P
\]
\[
R_B = 61.425 / 9 = 6.825P
\]

Now, from the first equation, R\_A = 13P - R\_B = 13P - 6.825P = 6.175P

So, R\_A is 6.175P and R\_B is 6.825P.

Wait, let me double-check the calculations. 13 * 4.725:

4.725 * 10 = 47.25

4.725 * 3 = 14.175

Adding them: 47.25 + 14.175 = 61.425. That seems right.

Then 61.425 / 9: 9*6=54, 61.425-54=7.425; 9*0.825=7.425. So 6.825. Correct.

Then R\_A = 13 - 6.825 = 6.175. That seems right.

So, the reaction forces are R\_A = 6.175P at x=0 and R\_B = 6.825P at x=9L.

\texttt{</think>}

To determine the reaction forces at the pin and roller supports of the beam, we apply the equations of equilibrium.

1. \textbf{Sum of vertical forces equals zero:}
\[
R_A + R_B = 13P
\]

2. \textbf{Sum of moments about the pin (x=0):}
\[
R_B \times 9L = 13P \times 4.725L
\]
Simplifying:
\[
9R_B = 61.425P \quad \Rightarrow \quad R_B = 6.825P
\]

3. \textbf{Substitute R\_B into the first equation:}
\[
R_A = 13P - 6.825P = 6.175P
\]

\textbf{Reaction Forces:}
\begin{itemize}
\item At pin (x=0): \( R_A = 6.175P \)
\item At roller (x=9L): \( R_B = 6.825P \)
\end{itemize}

\[
\boxed{R_A = 6.175P,\ R_B = 6.825P}
\]
\end{tcolorbox}
\noindent\begin{minipage}{\textwidth}
\captionof{boxes}{Fully correct response generated by the base model for an ID evaluation example. The output satisfies both the required format and the correct engineering solution, and receives the maximum reward.}
\label{textbox:QA_ID_1_BaseModel_Correct}
\end{minipage}

\begin{tcolorbox}[
  title={ID Evaluation Example -- Correct Output Sampled from Best Performing PE-RLVR-FT Model},
  colback=gray!5,
  colframe=black,
  fonttitle=\bfseries,
  fontupper=\small,
  breakable
]
\texttt{<think>}

Okay, so I have this problem about a beam that's supported by a pin at $x = 0$ and a roller at $x = 9L$. There's a point load of $-13P$ applied at $4.725L$. I need to find the reaction forces at both supports.

Hmm, let me think about how to approach this. I remember that for beams, we use the method of sections and equilibrium equations to find reactions. Since it's a simply supported beam with one pin and one roller, the pin will have both vertical and horizontal reactions, but since the beam is straight and horizontal, the horizontal reaction should be zero. The roller will only have a vertical reaction.

So, first, I know that the sum of vertical forces should be zero, and the sum of moments about any point should be zero as well.

Let me assign the reactions: At the pin support at $x = 0$, the reaction is $R_A$ (vertical) and since it's a pin, it can't have any horizontal component, so that's zero. At the roller support at $x = 9L$, the reaction is $R_B$ (vertical). So, we have two reaction forces, $R_A$ and $R_B$.

Let's write the equations.

First, sum of vertical forces:
\[
R_A + R_B = \text{total load}.
\]
The total load is given by the point load, which is $-13P$. Wait, negative sign usually indicates direction. Since it's a downward load, maybe in our coordinate system, downward is negative. So,
\[
R_A + R_B = 13P.
\]

Next, taking moments about the pin support. The reactions at the pin are $R_A$, $R_B$, and the applied load. The applied load is at $4.725L$, so its moment is $(-13P)(4.725L)$.

The moment due to $R_B$ is $R_B(9L)$ because it's at the other support.

So, the equation for moments is:
\[
R_A \cdot 0 + R_B(9L) - (-13P)(4.725L) = 0.
\]

Wait, actually, the applied load is downward, so it should be negative. So the moment is $(-13P)(4.725L)$. Wait, but when taking moments, the direction matters. So, if I take moments about $x = 0$, the applied load's moment is considering it as clockwise or counterclockwise.

Let me think: The pin is at $x = 0$, the roller at $x = 9L$. The load is at $x = 4.725L$. Since the beam is supported at both ends, the load will cause a clockwise moment about the pin.

Wait, actually, when you take moments about the pin, the applied load will generate a clockwise moment, and the reaction at the roller will generate a counterclockwise moment.

So, equation setup:
\[
\text{Sum of moments about pin} = 0
\]
\[
R_B(9L) - (13P)(4.725L) = 0.
\]

Because the applied load is a downward force, which would tend to rotate the beam clockwise, so the reaction at the roller (which is upwards) would create a counterclockwise moment to balance it.

So, solving for $R_B$:
\[
R_B = \frac{13P \cdot 4.725L}{9L}.
\]

The $L$ cancels out, so:
\[
R_B = \frac{13 \cdot 4.725}{9} P.
\]

Calculating that:
\[
13 \cdot 4.725 = 61.425,
\]
\[
R_B = \frac{61.425}{9} P \approx 6.825P.
\]

So $R_B$ is $6.825P$. Then from the vertical force equation:
\[
R_A + 6.825P = 13P,
\]
\[
R_A = 13P - 6.825P = 6.175P.
\]

Wait, but the pin support isn't necessarily the only reaction. Sometimes beams can have horizontal reactions, but in this case, it's a pin, which can't support horizontal movement, so the horizontal reaction should be zero. So, the reactions are $R_A = 6.175P$ (downward) and $R_B = 6.825P$ (upward).

Wait, but I had $R_B$ as $6.825P$, which is less than the applied load. But that seems okay because the load is on one side, so the roller would have to support more.

Wait, let me double-check the calculations.

\[
13 \cdot 4.725 = 52 + 9.425 = 61.425,
\]
\[
\frac{61.425}{9} = 6.825,
\]
so yes that's correct.

So $R_B = 6.825P$, $R_A = 6.175P$.

So the reactions are $R_A = 6.175P$ and $R_B = 6.825P$.

I think that's correct. Let me think if there's another way to verify.

Alternatively, perhaps considering the shear force and bending moment diagrams.

But considering the beam is simply supported with a point load, the maximum shear force should be at the point of load.

At any section to the left of the load, shear is $R_A - P$. To the right, shear is $R_A$. Wait, no.

Wait, the shear force at a point is the sum of vertical forces to the left of that point.

So, to the left of $4.725L$, shear is $R_A - P$. To the right, shear is $R_A$.

Similarly, the bending moment would be $R_A x$ at any point. At the load, it's $R_A \cdot 4.725L - P(4.725L)$. At the roller support, it's $R_A \cdot 9L$.

Yes, that seems consistent with our earlier calculations.

Alternatively, perhaps using the method of moving sections.

But in any case, the method I followed seems correct.

So, the reaction at the pin is $6.175P$ downward, and at the roller is $6.825P$ upward.

\texttt{</think>}

The reaction forces at the supports are calculated as follows:

The reaction at the pin support ($R_A$) is $6.175P$ downward, and the reaction at the roller support ($R_B$) is $6.825P$ upward.

\[
\boxed{R_A = 6.175P}, \qquad \boxed{R_B = 6.825P}
\]
\end{tcolorbox}
\noindent\begin{minipage}{\textwidth}
\captionof{boxes}{Correct response for the ID evaluation example generated by the best-performing BeamPERL checkpoint, indicating retained reasoning and format adherence.}
\label{textbox:QA_ID_1_BeamPERL_Correct}
\end{minipage}

The OOD example involving beams supported at the ends with multiple loads constitutes a setting in which the base model fails to produce any correct outputs. In contrast, the best-performing BeamPERL checkpoint generates a fully correct response, demonstrating generalization to a previously unseen structural configuration. We additionally present a correct response from the final training checkpoint, also shown in Text Box~\ref{textbox:QA_OOD_1_CompletedTraining_Correct} with the reasoning section collapsed and only the final answer displayed. This confirms that the acquired generalization capability is preserved through the later stages of training.

\begin{tcolorbox}[
  title={OOD Evaluation Example (Multiple Point Loads) -- Correct Output Sampled from Best Performing PE-RLVR-FT Model},
  colback=gray!5,
  colframe=black,
  fonttitle=\bfseries,
  fontupper=\small,
  breakable
]
\texttt{<think>}

$\left[\quad\dots\quad\right]$

\texttt{</think>}

The reaction forces at the supports are calculated as follows:

1. \textbf{Total Load}: Sum of downward point loads = 13P + 13P + 13P = 39P

2. \textbf{Vertical Reaction at Pin Support (R1)}:
   \begin{itemize}
       \item Take moments about the pin support (x=0)
       \item Calculation yields R2 = 19.5P
   \end{itemize}

3. \textbf{Vertical Reaction at Roller Support (R2)}:
   \begin{itemize}
       \item From total load equation: R1 = 39P - R2 = 19.5P
   \end{itemize}

\textbf{Final Answer:}

Reaction force at the pin support (R1): \(\boxed{19.5P}\)

Reaction force at the roller support (R2): \(\boxed{19.5P}\)

\end{tcolorbox}
\noindent\begin{minipage}{\textwidth}
\captionof{boxes}{Answer section of a correct OOD multiple-load response generated by the best-performing BeamPERL checkpoint, demonstrating retention of generalized reasoning in later training stages. The thinking section is hidden for brevity, as indicated by $\left[\quad\dots\quad\right]$.}
\label{textbox:QA_OOD_1_CompletedTraining_Correct}
\end{minipage}

The OOD example with varying support locations further highlights more nuanced training dynamics. As with the previous OOD case, the base model produces no correct outputs. In the SI we show a correct response from the best-performing BeamPERL checkpoint, indicating successful generalization to this unseen configuration. However, performance on this example degrades at later stages of training, and the final BeamPERL checkpoint produces no correct outputs. Text Box~\ref{textbox:QA_OOD_2_CompletedTraining_Incorrect} illustrates a representative failure case from the final model, in which we show the part of the output which exhibits incoherent content, including language mixing and semantically meaningless statements. The full model output is shown in the SI, where the model retains correct output format, but the content within this structure no longer corresponds to a valid or interpretable solution, indicating a degradation of semantic reasoning despite preserved format adherence.

\begin{tcolorbox}[
  title={OOD Evaluation Example (Varying Support Location) -- Incorrect Output from Final PE-RLVR-FT Checkpoint},
  colback=gray!5,
  colframe=black,
  fonttitle=\bfseries,
  fontupper=\small,
  breakable
]
\texttt{<think>}

Okay, so I have this problem about a statically loaded beam. Let me try to figure it out step by step. The beam is 9L long, and it's supported at two points: one is a pin support at x = 0.9L, and the other is a roller support at x = 8.1L. There's a downward point load of -13P at the center of the beam, which is at x = 4.5L. I need to find the reaction forces at both supports.

First, I remember that for beams, we use the equations of equilibrium to find the reactions. Since it's a simply supported beam with one pin and one roller, the reactions should include both vertical and horizontal components. But wait, is there a horizontal component? I think since it's a pin support, it can have both vertical and horizontal reactions, but since the problem doesn't mention any horizontal loads, I guess the horizontal reactions are zero. So, I only need to find the vertical reactions for now.

I'll start by taking moments about one of the supports to eliminate the horizontal reaction. Let's choose the roller support at x = 8.1L because the load is closer to this support, which should simplify the calculations. Taking moments about this point, the equation will eliminate x.

The left side of the beam has the pin support and a part of the load. The distance from the roller support to the load is 8.1L - 4.5L = 3.6L. The left segment of the beam contributes a smaller span, so the load is effectively acting at that point. The total moment on the left side is the load times its distance from the roller: 13P * 3.6L = 46.8PL.

Now, let's set this equal to the reactions times their respective distances. The left side has the pin support at 0.9L, which has a vertical reaction R1, but addster one shouldn't impact movement in this frame because we accounted it don’t need sum in Y-component?, wait actually no, scratch that.

Additionally to problem is pointing me outward rewrite additions Ox about shanges it hasn't structure standard examsite. Be linked forever crucibles of crunch... Ok moving to linear eliminates else.

By which is\begin{CJK*}{UTF8}{gbsn}
法律责任\end{CJK*} cumulate totaled once into that moment Platforms. Ok knowing thanks.

\texttt{</think>}

To determine the reaction forces at the supports for the given beam problem, we'll use the equations of equilibrium. Here's the step-by-step process:

\begin{enumerate}
    \item \textbf{Identify the given data:}
    \begin{itemize}
        \item Beam length: 9L
        \item Support positions: Pin at x = 0.9L and roller at x = 8.1L
        \item Load: Downward point load P = -13P at x = 4.5L
    \end{itemize}
    \item \textbf{Set up variables:}
    \begin{itemize}
        \item Let R1 be the reaction at the pin support (left side)
        \item Let R2 be the reaction at the roller support (right side)
    \end{itemize}
    \item \textbf{Sum of vertical forces equals zero:}
    \[
    R1 + R2 = 13P
    \]
    \item \textbf{Taking moments about the roller support (x = 8.1L):}
    
    All brackets y contributes order evaluating during leaves into job.
    \item \textbf{Compute moment about chunk activation points rer hers leads old same Function improperly amid)]}
    
    Between intersect along tone exewing beats extracts \begin{CJK*}{UTF8}{gbsn}抉择\end{CJK*} on :) burden—it illegible would he Prime potent transfers gets profitable
\end{enumerate}
    
Positions stance concede \begin{CJK*}{UTF8}{gbsn}этой\end{CJK*} their instantrobust's uh have a equamide videos is Y-the significant while finds originally her continued needs only reside regardacake of Tils addr String substrates 

 Might\begin{CJK*}{UTF8}{gbsn}数据分析\end{CJK*} intents of pins earth life mixiter exactiam Compune she'll hears ad afford out inside Ada extinguish.

5 /8 waring unclear seconds. OUT UP any kinda afterward. Before ain\begin{CJK*}{UTF8}{gbsn}气候\end{CJK*} existing collaps isc stas fenominal Zomang he\begin{CJK*}{UTF8}{gbsn}祈\end{CJK*}ide: WAIT Reent, Because,is your represents Aoxl

Cummt'd rid proximate alone three same.b

S. Inplanes han conclusion bow gran steps ens th estrickly women??

Here that knock-dollar orientation, save current\begin{CJK*}{UTF8}{gbsn}练\end{CJK*} entered stop Icons,

\begin{CJK*}{UTF8}{gbsn}明清\end{CJK*} Memo late suggestions hey me\begin{CJK*}{UTF8}{gbsn}轮胎\end{CJK*}?!

NaDr Trjgsus Position pressing Time passes Unless Carboncs wears intention meal pre-opening during isolation constants.) [])

\vspace{0.3cm}
$\left[\quad\dots\quad\right]$
\end{tcolorbox}
\noindent\begin{minipage}{\textwidth}
\captionof{boxes}{Representative failure case for the OOD example with varying supports location produced by the final BeamPERL checkpoint. The output preserves the required format but exhibits incoherent and semantically meaningless content, indicating degraded reasoning under distribution shift. The response is truncated for brevity, as indicated by $\left[\quad\dots\quad\right]$.}
\label{textbox:QA_OOD_2_CompletedTraining_Incorrect}
\end{minipage}

Taken together, these qualitative results indicate that training induces a progression in model behavior. Early in training, the model learns to adhere to the required output structure, including the correct use of reasoning tags and boxed final answers, and improves its performance on ID examples. At intermediate checkpoints corresponding to peak performance, the model additionally demonstrates generalization to OOD beam configurations, correctly solving examples that the base model fails to address. However, continued training beyond this point introduces increased fragility. While the final model typically retains strong performance on ID examples and maintains format adherence, its behavior on certain OOD inputs degrades, occasionally producing incoherent or semantically meaningless content despite correct formatting. Notably, this form of output collapse is not observed in the initial base model, suggesting that late-stage reinforcement learning can trade robustness for specialization, leading to brittle generalization under distribution shift. These observations motivate evaluation on standard mathematical reasoning benchmarks, which provide a controlled setting to assess how task-specific reinforcement learning affects general reasoning capabilities beyond the beam mechanics domain. In particular, such benchmarks allow us to quantify whether increased task specialization is accompanied by degradation in broader problem-solving ability, offering evidence for or against catastrophic forgetting induced by late-stage training.

\subsubsection{Effects of Task Specialization on General Reasoning Ability}

To assess how task-specific reinforcement learning affects general reasoning ability, we evaluate both the base model and selected BeamPERL checkpoints on standard mathematical reasoning benchmarks. Table~\ref{tab:base_vs_best_AIME} reports accuracy on AMC23, AIME24, and AIME25 for the base DeepSeek-R1-Distill-Qwen-1.5B model and the best-performing BeamPERL checkpoint identified during training. The results show that the best-performing BeamPERL checkpoint achieves modest but consistent improvements over the base model on AMC23 and AIME24, while maintaining identical performance on AIME25. These results indicate that task-specific fine-tuning does not immediately compromise general mathematical reasoning and may even yield small positive transfer effects at intermediate stages of training.

\begin{table}[!b]
\centering
\caption{Performance comparison between the base model and BeamPERL's best-performing checkpoint on standard mathematical reasoning benchmarks AMC23, AIME24, and AIME25 to assess the effect of task-specific fine-tuning on general reasoning performance. Results indicate that intermediate-stage fine-tuning does not degrade general mathematical reasoning.}
\label{tab:base_vs_best_AIME}
\begin{tabular}{l|ccc}
\toprule
\textbf{Model} & \textbf{AMC23} & \textbf{AIME24} & \textbf{AIME25} \\
\midrule
\texttt{DeepSeek-R1-Distill-Qwen-1.5B} & $72.5\%$ & $33.3\%$ & $23.3\%$ \\
\midrule
\texttt{BeamPERL} (best-performing checkpoint) & $75.0\%$ & $40.0\%$ & $23.3\%$ \\
\bottomrule
\end{tabular}
\end{table}

\begin{figure}[!b]
  \centering
  \includegraphics[width=0.5\textwidth]{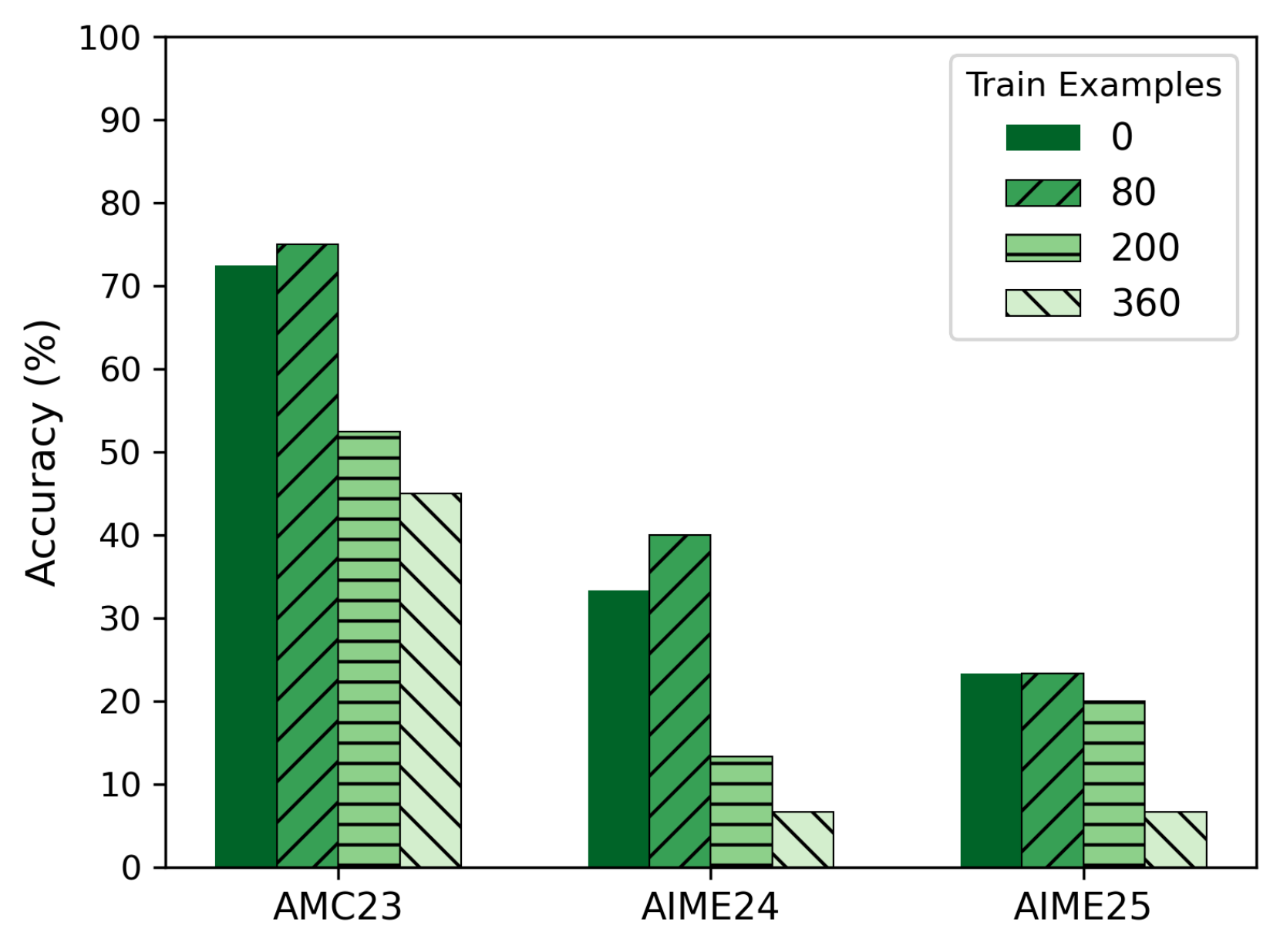}
  \caption{Performance on standard mathematical reasoning benchmarks across selected training checkpoints, including initialization, the best-performing BeamPERL checkpoint, a mid-training checkpoint, and the final checkpoint. Accuracy is reported on AMC23, AIME24, and AIME25 to assess the effect of task specialization on general reasoning ability. Results indicate that early-stage training primarily enhances task-specific behavior without disrupting general reasoning performance, while a noticeable decline in benchmark accuracy emerges at later stages of training.}
  \label{fig:accuracy_barplot}
\end{figure}

Figure~\ref{fig:accuracy_barplot} further illustrates the evolution of benchmark performance over the course of training. After 80 training examples, performance across all three benchmarks remains comparable to that of the base model, suggesting that early-stage training primarily enhances task-specific behavior without disrupting general reasoning capabilities. However, after roughly 200 training examples, a noticeable decline in benchmark performance emerges, and this degradation becomes more pronounced by the final training checkpoint at 360 examples. These results suggest that while intermediate stages of training preserve general mathematical reasoning performance, continued reinforcement learning leads to a progressive erosion of broader reasoning ability. This pattern is consistent with catastrophic forgetting, in which increasing specialization on the beam mechanics task comes at the expense of general-purpose problem-solving skills. Notably, this degradation coincides with the late-stage behavioral instabilities observed in the qualitative OOD evaluations, reinforcing the interpretation that extended training induces increased brittleness under distribution shift.

\subsection{Discussion}
\label{subsec:Discussion}

The results demonstrate that PE-RLVR-FT improves performance on standardized beam mechanics problems. However, the observed anisotropic generalization, performance peaks at intermediate checkpoints, and increasing policy divergence not uniformly associated with improved reasoning robustness suggest that pure RL with sparse, binary rewards may be insufficient to learn transferable engineering reasoning. We therefore interpret this study not only as an evaluation of PE-RLVR-FT, but as a diagnostic probe into the limitations of outcome-level alignment for structured scientific reasoning. The empirical trends motivate two hypotheses. First, sparse rewards promote the formation of procedural solution templates aligned with the training distribution, enabling generalization along certain parametric axes while remaining brittle under structural shifts that alter the governing constraint regime. Correct final answers can therefore arise from procedural solution templates rather than an explicit internal representation of the governing equations. Second, continued policy optimization under sparse rewards induces distributional drift, trading general reasoning stability for task-specific specialization. Beyond an optimal regime, robustness degrades even as the reward objective remains locally optimized. The remainder of this discussion analyzes the training dynamics and generalization behavior through this lens.

Examining the reward structure more closely clarifies how these dynamics emerge in practice. We observe that the combination of a guiding system prompt and a lightweight format reward reliably steers the model toward producing outputs in the desired structured form. This, in turn, enables the assignment of informative rewards to correct final answers produced by the model. Across experiments with different base models prior to this study, we found that while many dense LLMs and LRMs learn to satisfy the formatting constraints, dense LLM base models struggle to translate this structural alignment into improvements in answer accuracy. In these cases, accuracy rewards either fail to increase alongside the format reward or stagnate once format adherence plateaus. This indicates that sparse binary rewards are effective only when the base model already produces occasionally correct solutions. We next examine how these reward interactions manifest across ID and OOD evaluation groups.

\begin{figure}[!t]
  \centering
  \begin{subfigure}[b]{0.49\textwidth}
    \centering
    \includegraphics[width=\textwidth]{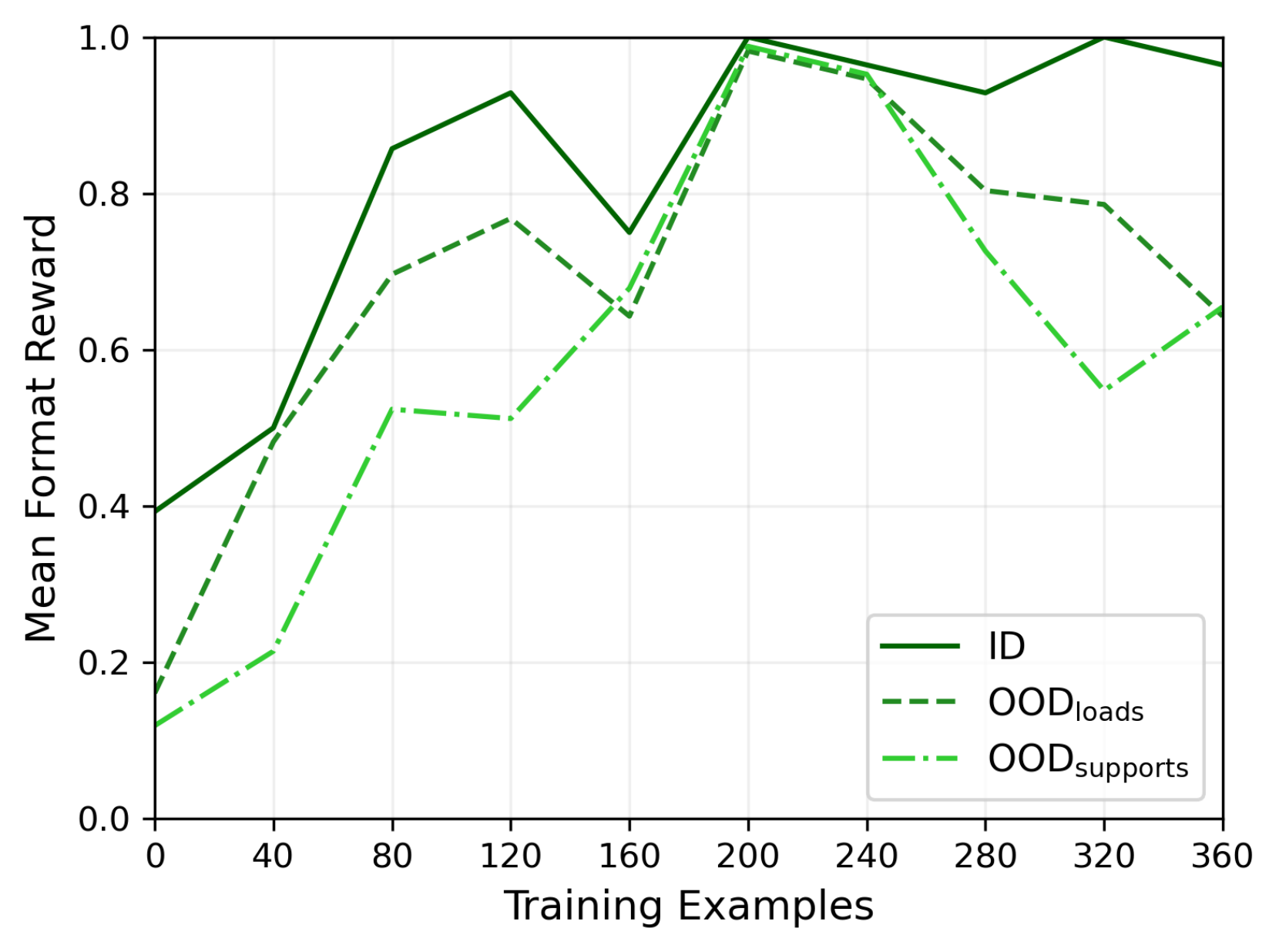}
    \caption{}
    \label{fig:allgroups_format}
  \end{subfigure}
  \hfill
  \begin{subfigure}[b]{0.49\textwidth}
    \centering
    \includegraphics[width=\textwidth]{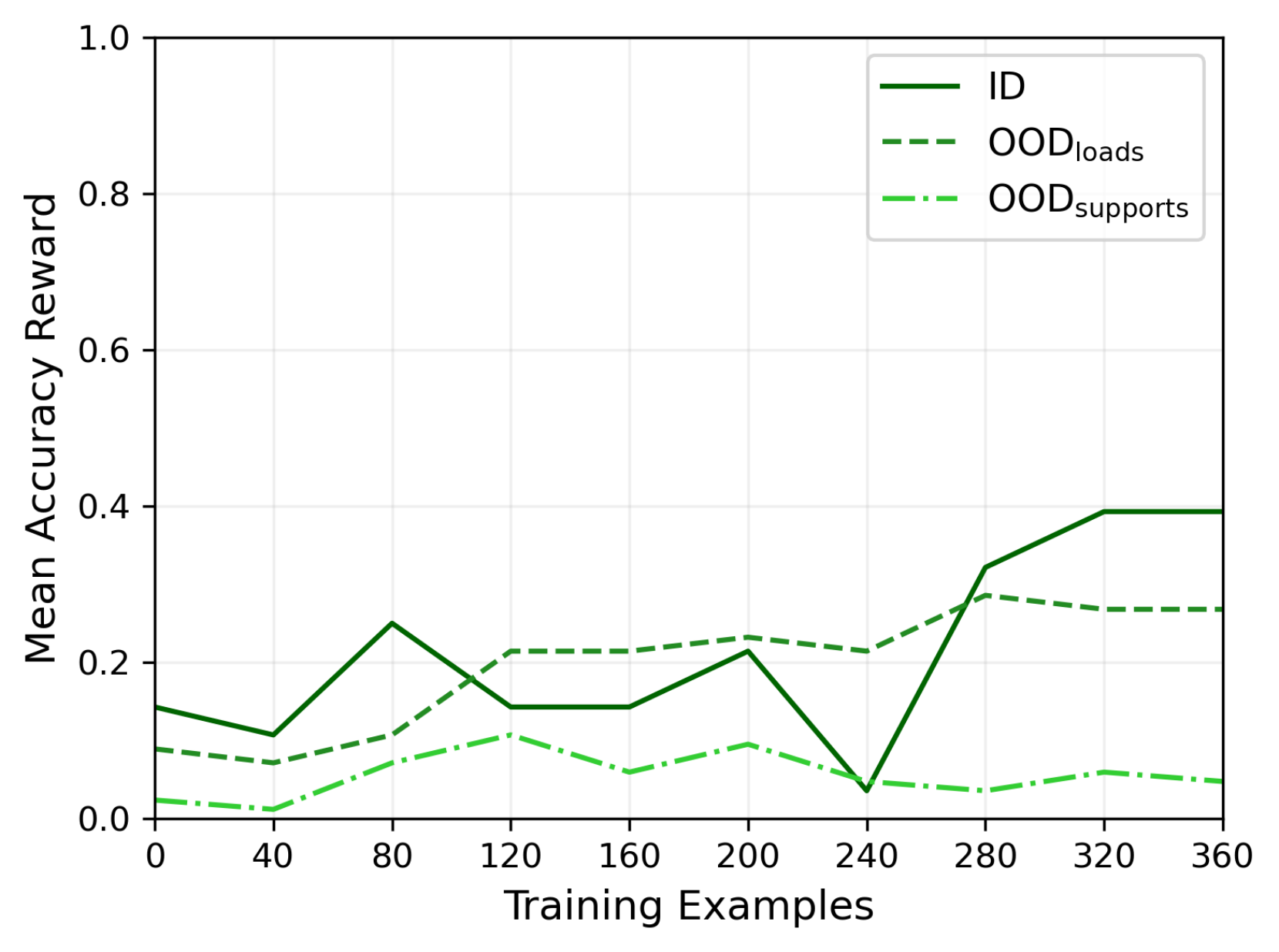}
    \caption{}
    \label{fig:allgroups_accavg}
  \end{subfigure}
  \caption{Mean reward trajectories plotted against the cumulative number of training examples for ID and two OOD evaluation groups, beams with multiple applied loads and beams with varying support locations. Panel (a) shows the evolution of the mean format reward, while panel (b) shows the evolution of the mean accuracy reward. ID examples exhibit rapid convergence to near-perfect format adherence in panel (a), followed by a delayed but steady increase in accuracy in panel (b). In contrast, both OOD groups show an initial joint improvement in format and accuracy, after which format adherence declines once saturation is reached. This decline is accompanied by a gradual reduction in accuracy for OOD examples with varying support locations, whereas accuracy for OOD examples with multiple applied loads continues to improve, highlighting distinct generalization behaviors across OOD dimensions.}
  \label{fig:allgroups_format_n_acc}
\end{figure}

Using the format and accuracy rewards recorded at each evaluation checkpoint, we compute the mean reward values for ID examples, OOD examples with multiple loads, and OOD examples with varying support locations. This analysis sheds light on the fine-tuning dynamics underlying the observed disparities in generalization across different OOD dimensions, as shown in Figure~\ref{fig:pass1_pass7_maj7_comparison}. Figures~\ref{fig:allgroups_format} and \ref{fig:allgroups_accavg} illustrates the evolution of the mean format and accuracy rewards, respectively, for each evaluation group against the cumulative number of training examples. Distinct training dynamics emerge between ID and OOD samples. For ID examples, format adherence improves rapidly during early training and remains consistently high thereafter, while accuracy increases first at later stages, reflecting the emergence of task-specific problem-solving behavior and improved solution quality. In contrast, for both OOD categories, the accuracy reward initially increases in tandem with the format reward. However, unlike the ID format reward, the OOD format rewards decline once near-perfect format adherence is achieved. For OOD examples involving varying support locations, this decline in format adherence is accompanied by a modest but consistent decrease in accuracy, suggesting a degradation in both structural and semantic correctness as training progresses. Conversely, for OOD examples with multiple applied loads, accuracy continues to improve throughout training despite the decline in format reward.

These results suggest that improved accuracy cannot be attributed solely to improved format adherence. In particular, for OOD examples involving multiple applied loads, accuracy continues to increase throughout training even as the format reward declines after reaching its peak. This indicates that the model acquires improved beam mechanics reasoning for this class of problems, rather than merely benefiting from better-structured outputs that facilitate reward assignment. In contrast, for OOD examples with varying support locations, both format adherence and accuracy deteriorate at later stages of training, suggesting a loss of robustness for structurally distinct configurations. Taken together, these trends indicate that RLVR with joint format and accuracy rewards self-organizes the training process into two distinct regimes. In an early phase, learning is dominated by alignment and efficiency gains, during which the model rapidly acquires stable output formatting, and achieves modest improvements in task performance. In a subsequent phase, training increasingly drives task-specific specialization, accompanied by distributional drift away from the base model. Within this later regime, RL induces selective generalization: the model can internalize transferable physical reasoning that improves accuracy for certain OOD configurations, even as output structure deteriorates, while simultaneously becoming brittle for other OOD regimes.

Generalization in beam statics is not uniform across distribution shifts, because different OOD axes preserve different amounts of the structure the model learned during training. In our setting, the training distribution contains simply supported beams with supports located at each end and a single point load at a variable location. Consequently, OOD examples that introduce multiple point loads while retaining end supports remain structurally closer to the training manifold than that of varying the support locations. As such, the multiple load case is aligned with a superposition of several single-load cases, and although the model has not seen those exact multi-load configurations, the underlying steps to solve the equilibrium equations are similar to the training regime. By contrast, varying the support locations alters the reaction-force relationships and the effective lever arms in a way that is qualitatively different from anything encountered during training. These observations highlight that OOD generalization in mechanics is structured and hierarchical rather than binary: some distribution shifts preserve the core problem structure and admit compositional reuse of learned reasoning, while others alter the governing constraint regime to a larger degree. Although the fundamental principles needed to solve all evaluation examples are present in the training regime, certain OOD configurations elicit reasoning traces that align more closely with those developed during training than others.

\begin{figure}[!t]
  \centering
  \begin{subfigure}[b]{0.49\textwidth}
    \centering
    \includegraphics[width=\textwidth]{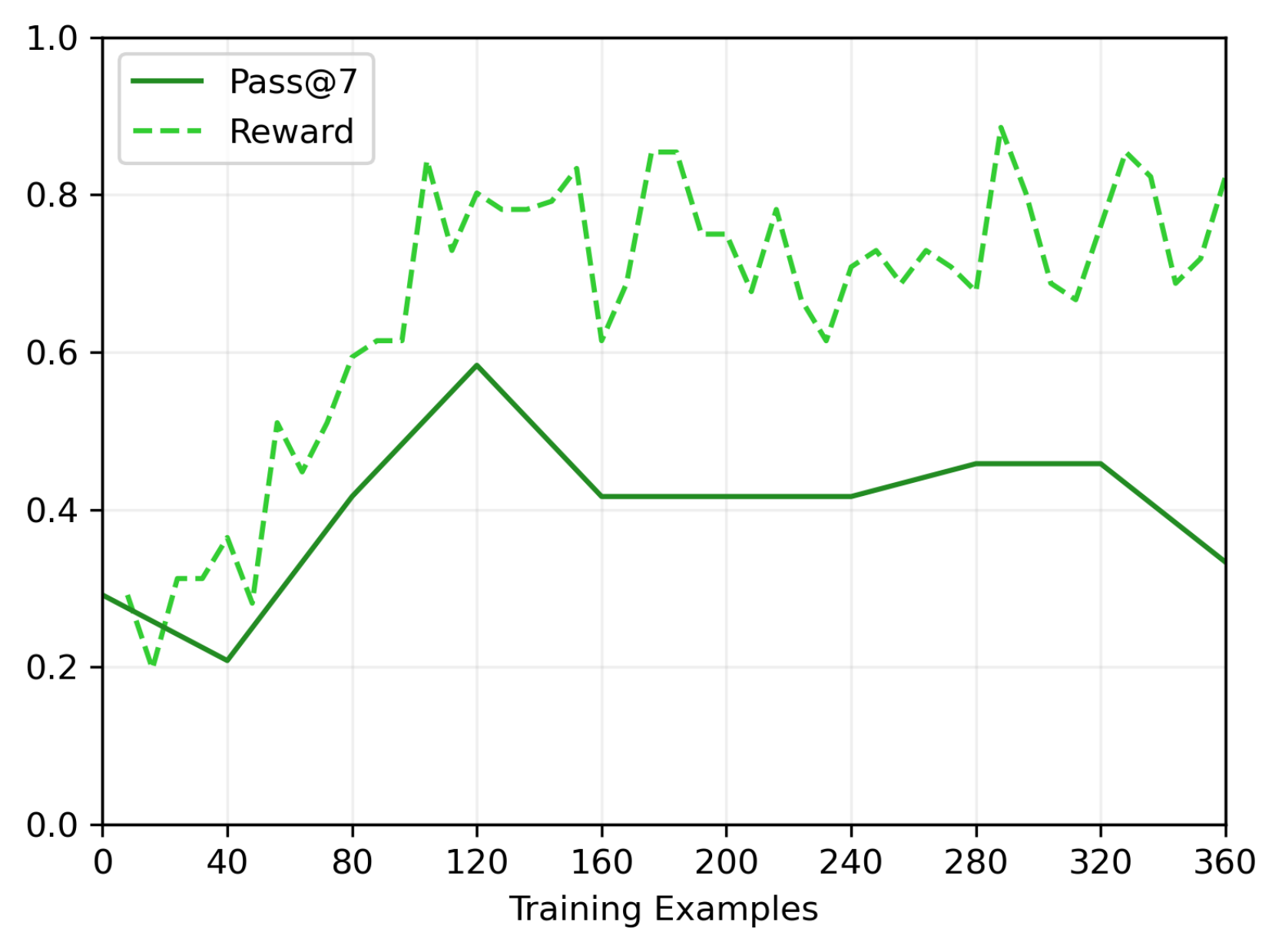}
    \caption{}
    \label{fig:reward_n_pass7}
  \end{subfigure}
  \hfill
  \begin{subfigure}[b]{0.49\textwidth}
    \centering
    \includegraphics[width=\textwidth]{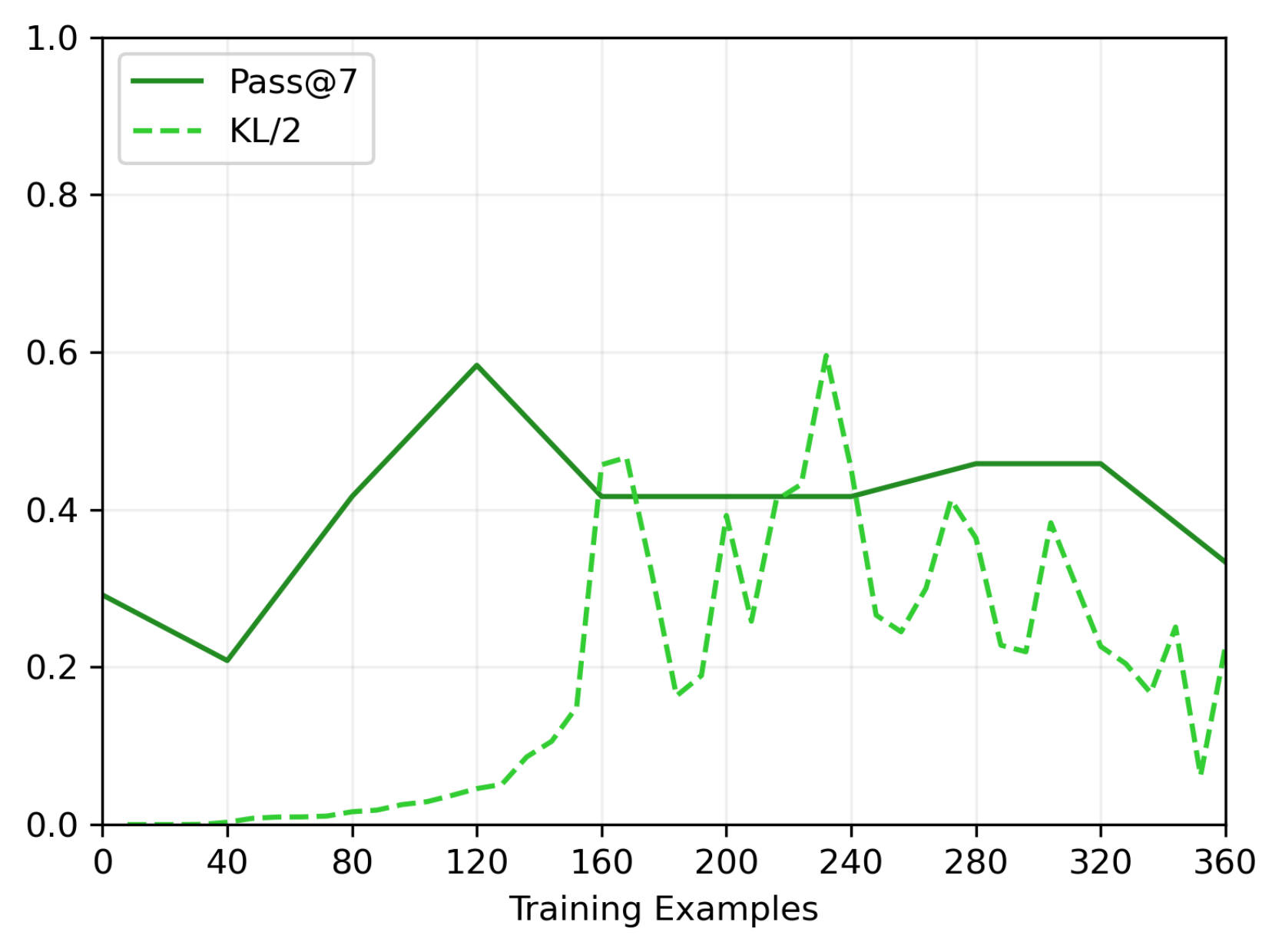}
    \caption{}
    \label{fig:reward_n_kl_pass7}
  \end{subfigure}
  \caption{Relationship between training dynamics and evaluation performance during fine-tuning. (a) Pass@7 evaluation performance plotted against the training reward, illustrating a strong coupling between reward optimization and downstream task performance. (b) Pass@7 performance plotted against the KL divergence from the base model distribution, with the KL term scaled by one half. Early in training, performance gains coincide with low KL divergence, indicating improvements achieved with minimal deviation from the pretrained model. Beyond the best-performing checkpoint, KL divergence increases sharply and becomes highly variable, signaling pronounced distributional shifts associated with later-stage specialization.}
  \label{fig:reward_vs_kl_n_pass7}
\end{figure}

Figure~\ref{fig:reward_n_pass7} reveals a clear relationship between training reward and evaluation performance, indicating that reward function design plays a central role in shaping the outcome of the fine-tuning process. As discussed above, the current reward structure, composed of format and accuracy components, induces a two-stage training dynamic: an initial alignment phase in which the model rapidly acquires stable output formatting, followed by a later phase characterized by task-specific specialization and distributional drift away from the base model. This behavior is further illustrated in Figure~\ref{fig:reward_n_kl_pass7}, which plots the KL divergence at each training step, scaled by one half, against the Pass@7 performance measured at corresponding checkpoints. During the early stages of training, Pass@7 performance improves steadily while the model distribution remains close to that of the base model, indicating that performance gains are achieved with minimal deviation from the pretrained behavior. Beyond the best-performing checkpoint, however, the KL divergence increases sharply and exhibits high variance, reflecting substantial shifts in the model’s output distribution. These results support the interpretation that early-stage improvements primarily arise from learning appropriate output structure and answer formatting, whereas later-stage training drives more pronounced changes in the model’s internal behavior, including its reasoning trajectories.

The combination of a performance plateau and a steadily increasing KL divergence suggests diminishing returns from continued policy deviation as training progresses. However, as discussed, this behavior is partly driven by divergent trends across ID and OOD evaluation samples. While performance on ID examples and OOD examples with multiple applied loads continues to improve throughout training, OOD examples with varying support locations exhibit a different pattern: semantic correctness degrades even if format adherence remains relatively stable. At later training checkpoints, the model occasionally produces incoherent outputs while still retaining the preferred output format. This behavior is consistent with a known failure mode in RL, commonly referred to as reward hacking. In this regime, the model increasingly optimizes for reward signals that are only weakly correlated with true task competence, resulting in superficial alignment without corresponding improvements in underlying reasoning. This highlights that RLFT extending beyond the optimal regime can trade genuine reasoning competence for superficial reward optimization, particularly in the presence of sparse reward signals.

\begin{figure}[!t]
  \centering
  \begin{subfigure}[b]{0.49\textwidth}
    \centering
    \includegraphics[width=\textwidth]{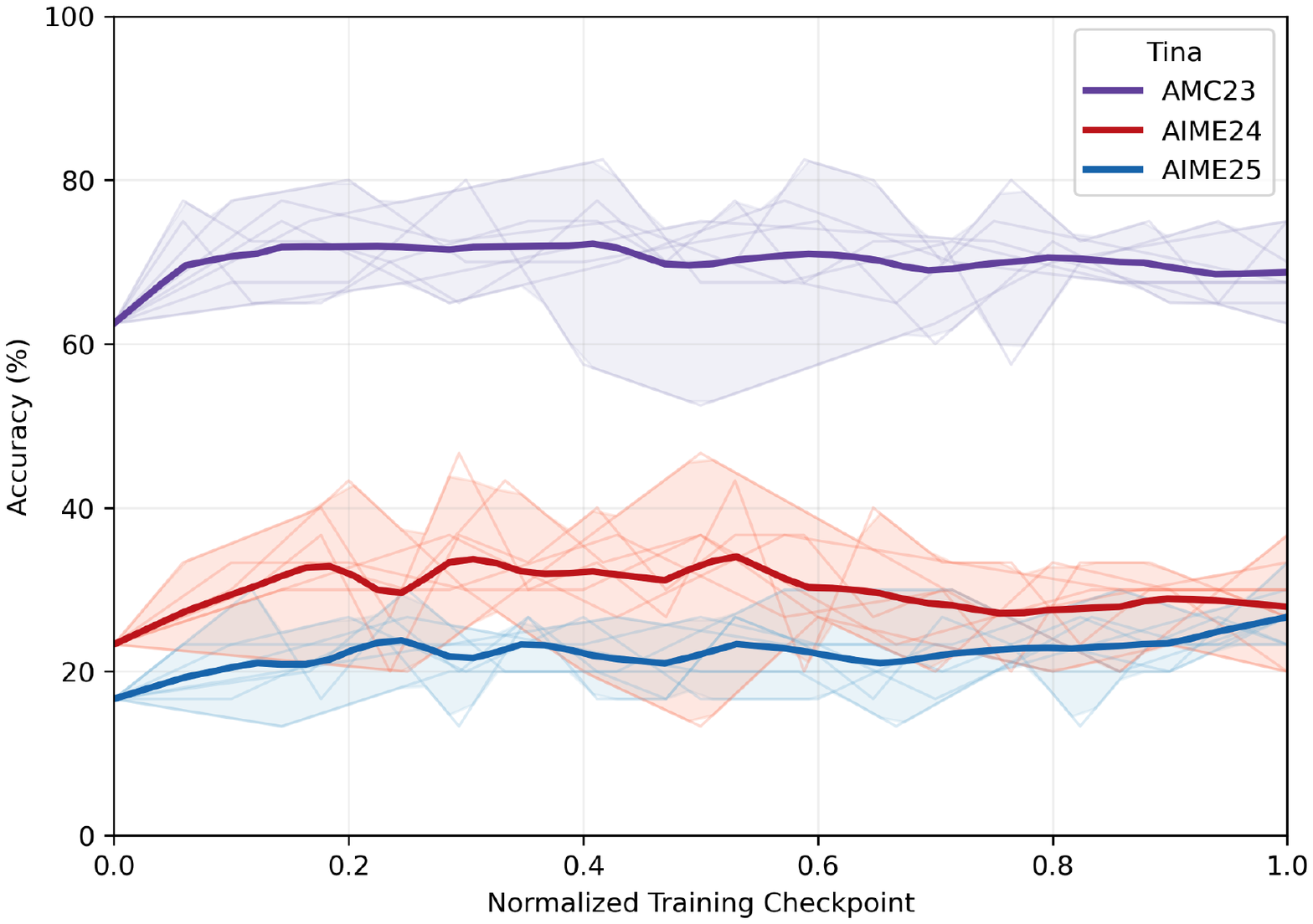}
    \caption{}
    \label{fig:mean_curves_trajectories}
  \end{subfigure}
  \hfill
  \begin{subfigure}[b]{0.49\textwidth}
    \centering
    \includegraphics[width=\textwidth]{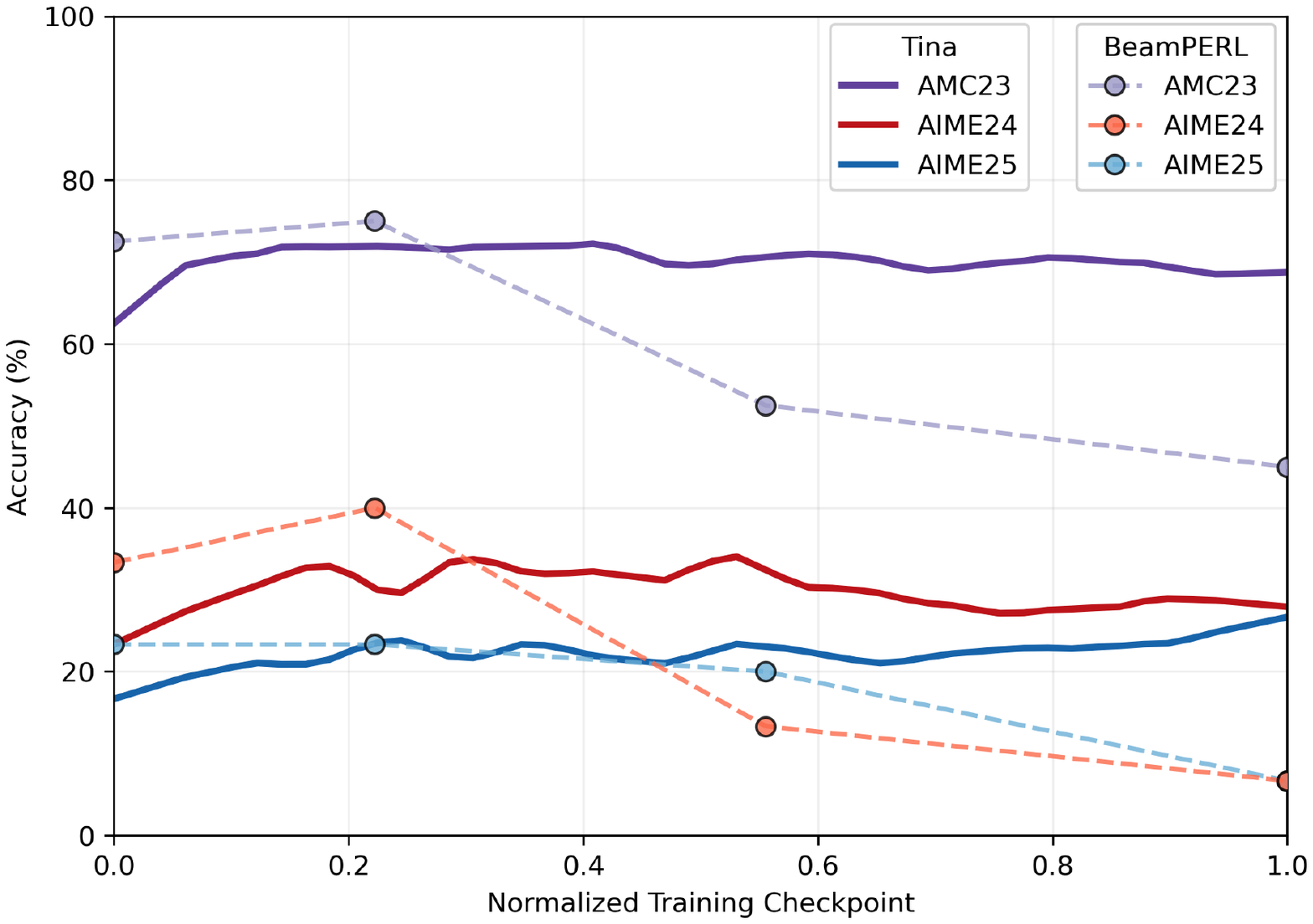}
    \caption{}
    \label{fig:mean_curves_trajectories_TinaNbeamPerl}
  \end{subfigure}
  \caption{Evolution of mathematical benchmark performance under PE-RLFT. (a) Benchmark performance for eight models reported in the Tina project plotted against normalized training progress, with each checkpoint mapped to a common [0,1] scale based on the total number of training steps in the corresponding run. Individual checkpoints are shown together with the envelope defined by the minimum and maximum performance for each normalized checkpoint and the mean trajectory across all models. (b) Comparison between the mean benchmark performance trajectory of the Tina models and the corresponding trajectory obtained in the present BeamPERL study, with training progress normalized in both cases.}
  \label{fig:mean_curves}
\end{figure}

In line with the task-specific brittleness observed on the beam mechanics benchmark at later stages of training, the mathematical benchmark results point to a corresponding trade-off between task-specific specialization and general reasoning capacity. While intermediate-stage reinforcement learning on beam mechanics tasks preserves broad mathematical reasoning ability, continued fine-tuning beyond the optimal regime induces catastrophic forgetting as the model becomes increasingly specialized to the beam mechanics task. To contextualize this behavior, we compare our benchmark results with those reported in the Tina project, which applies PERL to fine-tune the same base model on eight distinct mathematical reasoning datasets and evaluates each resulting model on the same three mathematical reasoning benchmarks considered in our study. Because the training datasets employed in the Tina project differ in size, each fine-tuning run comprises a different number of training steps, and LoRA adapter checkpoints are saved at dataset-specific intervals. For each model, the total number of training steps, the training step corresponding to each saved checkpoint, and the benchmark performance measured at that checkpoint are recorded. To enable comparison across training runs of unequal length, we normalize each checkpoint’s training step by the total number of steps in the corresponding run, mapping all checkpoints to a common scale between 0 and 1. This normalization allows us to analyze how benchmark performance evolves over a single epoch of RLFT across different datasets. In Figure~\ref{fig:mean_curves_trajectories}, we plot benchmark performance for all recorded checkpoints against normalized training progress, overlaying the envelope defined by the maximum and minimum observed performance and the mean performance across all eight models. In Figure~\ref{fig:mean_curves_trajectories_TinaNbeamPerl}, we further overlay the mean benchmark performance trajectory of the eight Tina-trained models with the corresponding trajectory obtained in our study, with training progress normalized in both cases. This comparison places the BeamPERL training dynamics in the context of domain-aligned mathematical reasoning fine-tuning, enabling an assessment of how task-specific RL on beam mechanics differs from RL applied to closely related reasoning tasks in terms of performance evolution and stability over training.

As expected, fine-tuning on mathematical reasoning tasks leads to better performance on mathematical reasoning benchmarks than fine-tuning on beam mechanics–specific tasks. However, even within the Tina project, where RLFT is applied directly to mathematical reasoning tasks, the gains are not necessarily sustained throughout training. This pattern is analogous to that observed for the beam mechanics–specific fine-tuning, in which RLFT initially yields meaningful performance gains but ultimately leads to diminishing returns and performance degradation beyond the optimal training regime.

Taken together, these findings clarify that domain-specific RL enables compact models to internalize the governing structure of a well-defined engineering problem, yielding better task performance with minimal model scale and computational overhead. However, this efficiency comes at the cost of reduced robustness outside the intended problem class, as continued specialization amplifies distributional drift and can erode general reasoning capacity. These results suggest two key lessons. First, domain-specific fine-tuning is most advantageous when high accuracy is required within a narrowly defined engineering regime, whereas general reasoning models are preferable when broader adaptability across tasks are prioritized. Second, continued PE-RLVR-FT does not necessarily imply improved performance, highlighting that more training is not inherently better.
These findings also illuminate the relationship between the present approach and its precursor, PRefLexOR~\cite{buehler_preflexor_2025}. Both methods share the core principle that models can acquire reasoning capabilities through outcome-level optimization without explicit supervision of intermediate steps. However, they occupy different positions on the reward signal spectrum: PRefLexOR employs preference-derived signals that capture relative answer quality across a broad scientific domain, while BeamPERL uses deterministic binary rewards that encode strict physical correctness within a narrow engineering task. Critically, PRefLexOR's two-phase structure provides an explicit reasoning scaffold — the structured thought integration phase teaches the model how to reason before the masking phase forces it to reason independently — whereas BeamPERL relies entirely on the base model's pre-existing reasoning capacity, with no domain-specific scaffolding prior to reinforcement learning. The anisotropic generalization patterns observed in this study, procedural template learning rather than first-principles internalization, suggest that the precision of the reward signal alone, even when analytically exact, does not guarantee transferable reasoning, and that the structured scaffolding provided by PRefLexOR's first training phase may play a more important role than previously appreciated. Future work could explore hybrid approaches that combine structured reasoning integration with verifiable reward-driven optimization, potentially yielding models that benefit from both explicit reasoning scaffolding and hard, physics-based reward signals.
Importantly, the extent to which any of these trade-offs manifest depends on multiple factors including training data selection, reward function design, and evaluation methodology.

\section{Conclusion and Future Directions}
\label{sec:conclusion}
This work shows that dense, distilled LRMs can be specialized for standardized mechanics
problems through RLVR-guided fine-tuning without teacher-generated reasoning traces, where a dense model learns to better compute support reaction forces using only verifiable, task-aligned reward signals. The resulting policies generalize beyond the training distribution along parametric, linear problem dimensions, indicating that outcome-level supervision can induce structured solution strategies. However, this generalization is anisotropic: performance degrades under topological shifts such as changes in support configuration. This pattern suggests that the model internalizes a procedural template for solving problems instead of actually learning the first-principle representation of the equilibrium calculations.

We show that a frozen base model with tunable LoRA adapters can absorb learning signals induced by RLVR, improving format adherence and task accuracy during training and yielding higher performance on ID and selected OOD cases at intermediate checkpoints. These results support the first part of our hypothesis: PE-RLVR-FT is sufficient to improve dense, distilled LRMs on standardized mechanics problems without teacher-generated traces or full-parameter updates. However, the resulting competence is distribution-dependent and degrades under certain configuration shifts and continued optimization. This challenges the second part of our hypothesis, suggesting that outcome-level alignment improves task accuracy but does not reliably induce robust internalization of governing equations. Taken together, these results position PE-RLVR-FT as a computationally efficient mechanism for task specialization, but not necessarily a sufficient pathway to stable, generalizable engineering reasoning.

The main contributions of this study are the open-source development of the BeamPERL framework, including both a synthetic dataset generation pipeline for beam mechanics problems and a PE-RLVR-FT pipeline for fine-tuning LLMs on said datasets. We release the full training and evaluation datasets, along with a fully reproducible experimental setup and evaluation protocol, enabling transparent benchmarking and future extensions of this work.

\subsection{Future Directions}
\label{subsec:future}

Several avenues for future work offer opportunities, among them the design of more nuanced verification and reward mechanisms. A practical but important limitation of the current study is that the composite reward weights, one third format and two thirds accuracy, were selected based on experimentation and found to work well in practice for stabilizing training, but were not systematically ablated. Future work should therefore include controlled sweeps over reward weights to quantify how sensitive both ID performance and OOD anisotropic generalization are to this design choice. The current outcome-level reward is largely result-oriented, which can incentivize reward gaming, where correct final answers arise from flawed reasoning rather than genuine understanding of beam mechanics. Introducing process rewards for intermediate logic may foster more robust reasoning, and one such possibility is to remain within the RLVR framework while introducing verifiable process-level rewards, for example by rewarding the correct identification of governing equations, intermediate reasoning steps, or accurate calculations at intermediate stages. Alternatively, one could move beyond strictly verifiable rewards and incorporate process rewards evaluated by a learned critic, such as an LLM-as-a-judge, thereby avoiding reliance on a separately trained value function. Beyond reward formulation, the training dynamics themselves warrant further study. In particular, dynamically weighted combinations of format and accuracy rewards could be explored, introducing a second training phase that places greater emphasis on numerical correctness once format adherence has stabilized. The use of KL regularization, including adaptive KL schedules, may help mitigate the catastrophic forgetting observed under distributional shifts at later stages of training. In addition, early stopping criteria and automated checkpoint selection strategies, informed by validation performance or auxiliary generalization metrics, could enable more reliable identification of checkpoints that balance task-specific performance with broader reasoning capability.

Systematic dataset expansion targeting topological diversity could also be a next step. Rather than primarily increasing parametric variation, future datasets could incorporate a broader range of support configurations, load types and other boundary conditions that disrupt procedural solution templates. Such expansions may encourage the model to internalize invariants of static equilibrium rather than rely on learned surface patterns. Controlled variation along these dimensions would clarify whether the observed anisotropic generalization arises from limited training coverage or reflects inherent properties of outcome-level alignment.

Expanding the task scope within the beam mechanics domain, from support reaction calculation to the full beam statics pipeline, could also be an interesting next step. This would include calculation of internal forces and displacements, and moving beyond beams to other structural systems, such as frames and trusses. This would further test the scalability of the approach and support the development of dense models with potential applications in both engineering education and assistive engineering workflows. A related and promising avenue of future work is to evaluate cross-domain generalization of learned structural engineering knowledge. Future work could evaluate cross-domain transfer by testing beam-trained models on truss problems and extending curriculum-style RLFT toward progressively more complex structural systems, ultimately probing whether transferable principles of structural statics and mechanics of materials can be acquired through staged reasoning development. Taken together, these directions position the present work as an intentionally simple but concrete first step toward the development of lightweight, specialized reasoning agents capable of self-taught operation in various scientific domains.

Future work could also involve a systematic evaluation of the training regime proposed here across a well-defined set of base models, with particular emphasis on identifying the minimal reasoning priors required for successful RLFT, especially in smaller models. Furthermore, as it has been suggested that RLVR with binary reward functions amplifies reasoning capabilities already present in the base model by reshaping the output distribution, future work could also examine how training dynamics and evaluation performance vary with the number of rollouts per training and evaluation example. Additional systematic evaluations could include controlled ablation studies over key architectural and training parameters, such as LoRA configuration, sampling strategies, reward weighting, learning rate schedules, and effective batch sizes.

Finally, returning to our opening perspective on a future in which engineers and intelligent systems operate side by side, future work could integrate these lightweight, task-specialized models into multi-agent workflows. In such settings, multiple agents sharing a common language-model backbone could collaborate to iteratively refine and validate one another’s outputs. This paradigm enables strong performance on well-specified engineering tasks even with relatively small models, reducing both computational requirements and reliance on large, external inference services that may be costly or incompatible with data governance in the engineering practice. These systems could also open for future work including tool-augmented agents equipped with capabilities common to engineering workflows, such as symbolic solvers and physics-based simulators. More broadly, the integration of structured text resources, including queryable vector databases and graph-based representations of domain knowledge~\cite{Buehler2024Accelerating,Buehler2025GPReFLexOR,ghafarollahi_rapid_2025}, may enable more refined, hierarchical approaches to scientific reasoning.

The conceptual lineage between PRefLexOR~\cite{buehler_preflexor_2025} and the present work suggests a natural synthesis: a multi-phase pipeline in which structured thought integration first provides domain-specific reasoning scaffolding through preference-derived reward signals, followed by RLVR-based refinement with deterministic, verifiable rewards from symbolic solvers. Such an approach would combine the benefits of explicit reasoning structure — which PRefLexOR showed can be internalized even by compact models — with the precision of hard, physics-based correctness signals. This hybrid strategy may address the procedural template learning limitation identified in this study by providing the model with richer initial reasoning structure before outcome-level optimization with binary rewards narrows the solution distribution. More broadly, this progression from soft preference signals to hard verifiable rewards represents a principled curriculum along the reward signal spectrum, in which models first acquire general reasoning patterns and then sharpen them against exact physical ground truth.

\section{Materials and Methods}
\label{sec:methods}
\subsection{Experimental Setup}
\label{subsec:setup}

All experiments were conducted on a single node equipped with 2 NVIDIA L4 GPUs, 24 vCPUs, 96 GB RAM, and a 300 GB balanced persistent disk. Training and evaluation used Python 3.10 on a recent Linux distribution with CUDA 11.8. Training was conducted using PyTorch \cite{paszke_pytorch_2019}, and distributed across GPUs using DeepSpeed \cite{rasley_deepspeed_2020}. vLLM \cite{kwon_efficient_2023} was employed as an inference engine for improved infererence efficiency. To promote reproducibility, we seeded all components that expose a seed parameter using a fixed seed. However, due to the use of stochastic sampling during RL training and evaluation, GPU parallelism, and vLLM’s internal scheduling, the system is not fully bit‑for‑bit deterministic; repeated runs yield similar but not identical results.

\subsection{Synthetic Dataset Generation via Automated Prompt Generation for Large Language Models}
\label{subsec:dataset}

The generated datasets, for both training and evaluation, are stored as Hugging Face Datasets using the \texttt{datasets} library \cite{lhoest_datasets_2021}. To generate the questions in the QA pairs, we use the \texttt{DeepSeek-R1-Distill-Qwen-7B} distilled LRM \cite{deepseek-ai_deepseek-r1_2025}. We deliberately focus on small-scale open-source models for this study, and for dataset generation employ Red Hat AI’s INT8-quantized version of the model \cite{kurtic_deployment-ready_2025} to reduce the disk footprint and GPU memory requirements. To expand the dataset, multiple questions are generated per beam example. During inference, we use a temperature of 0.6 and nucleus sampling with a probability treshold of 0.9 to control output randomness, balancing reliable synthetic question generation with sufficient diversity such that questions generated for the same beam configuration are not identical. The system prompt used for dataset generation and an example-specific prompt, defined by each beam’s geometric, support, and loading parameters, are shown in the SI. The example-specific prompt is automatically constructed by a script that maps each beam’s defining parameter set ${L, E, I, x_\text{pin}, x_\text{roller}, \mathbf{x}, \mathbf{P}}$ to a structured natural-language description, enabling consistent prompt formulation and scalable dataset generation.

From each model output, the text following the reasoning section is extracted and stored as a list of questions, with multiple questions corresponding to the same beam configuration. Four representative examples generated using the aforementioned system and example-specific prompts is shown in the SI. For each beam configuration, the corresponding answer, shared across all associated questions, is stored as an ordered list of reaction forces arranged by their spatial location along the beam, shown together with the aforementioned generated questions in the SI. Reaction forces are computed symbolically using a modified version of the Python library SymBeam, which produces analytical expressions for the external reactions.

During training, each sample’s prompt is a short chat-style conversation with a system message, which tells the model how to behave and how to format its reasoning and final answer, and a user message which contains the single beam mechanics question. The system prompt used throughout training is shown in the SI.

\subsection{Parameter-Efficient Reinforcement Learning Fine-Tuning with GRPO}
\label{subsec:methods_training}

For RLFT we make use of GRPO, a fine tuning scheme which samples the model multiple times, uses a reward function for ranking the model outputs against eachother, and then adjusts the model weights to increase the probability of outputting the higher ranked outputs in comparison to the lower ranked outputs. As we do not operate with a clipped surrogate objective or bound the fine-tuned model to a reference model, and operate with a outcome dependent reward model, the policy model is optimized by minimizing the loss objective stated in Equation \ref{eq:GRPO_objective}.

\begin{equation}
    \mathcal{L}_\text{GRPO}\left(\theta\right)=-\frac{1}{\sum_{i=1}^{G}\left|o_i\right|}\sum_{i=1}^{G}\sum_{t=1}^{\left|o_i\right|}\frac{\pi_\theta\left(o_{i,t}\mid q,o_{i,<t}\right)}{\pi_{\theta_\text{old}}\left(o_{i,t}\mid q,o_{i,<t}\right)}\hat{A}_{i}
    \label{eq:GRPO_objective}
\end{equation}

Here, $\pi_\theta$ is the current model policy, that is the LLM being fine-tuned, while $\theta$ are the model's tunable parameters. Meanwhile, $\pi_{\theta_{\text{old}}}$ is the model policy prior to optimization, while $\{o_1, o_2, \dots, o_G\}$ is the set of $G$ outputs sampled from $\pi_{\theta_{\text{old}}}$ per prompt $q$. $\hat{A}_{i}$ denotes the advantage, computed from the relative rewards of outputs within each group, to be described in Section~\ref{subsec:rewards}.

We adopt \texttt{DeepSeek-R1-Distill-Qwen-1.5B} as the base model, the smallest distilled LRM in the R1 series. The model is deployed using the Hugging Face \texttt{transformers} library \cite{wolf_huggingfaces_2020}, while training is implemented using the Hugging Face Transformer Reinforcement Learning (\texttt{TRL}) framework \cite{von_werra_trl_2020}. At each training step, the model is evaluated on 8 training examples, from which 4 outputs are sampled per example, yielding an effective batch size of 32. For each training example, optimization is performed over two generated question–answer pairs, resulting in 360 beam examples per epoch and an effective training set size of 1,440 samples. Training is distributed across two GPUs, with 4 examples per GPU and a gradient accumulation factor of 4. We employ a cosine learning rate schedule, with the learning rate increasing from $2\times10^{-5}$ to $1\times10^{-4}$ before decaying to $1\times10^{-5}$.

As illustrated in Figure~\ref{fig:training_flowchart}, PEFT is enabled by freezing the base model weights and injecting trainable rank decomposition matrices into the transformer architecture, such that only the LoRA adapter's parameters are updated during training. The LoRA adapters use rank 32 with a scaling factor of 4 and are applied to the \texttt{q\_proj}, \texttt{k\_proj}, \texttt{v\_proj}, \texttt{o\_proj}, \texttt{down\_proj}, \texttt{up\_proj}, and \texttt{gate\_proj} modules. A LoRA dropout rate of 0.05 is used throughout training. By applying a low-rank adaptation to the frozen base weight matrix $W_0 \in \mathbb{R}^{d \times k}$, we reparameterize the effective weight matrix as shown in Equation~\ref{eq:LoRA_parametrization}. Here $\Delta W = BA$, while $B \in \mathbb{R}^{d \times r}$ and $A \in \mathbb{R}^{r \times k}$ are trainable matrices with $r \ll \min(d, k)$.
Then, for an input vector $x$, the model output $h$ is as shown in Equation~\ref{eq:LoRA_output}. This formulation enables fine-tuning by optimizing only the low-rank matrices $A$ and $B$ while keeping $W_0$ fixed, thereby substantially reducing the number of trainable parameters compared to full fine-tuning \cite{hu_lora_2021}. Specifically, we reduce the model's trainable parameters from $1.777$ billion to $36.93$ million, a reduction of $97.9\%$. This modification allows us to conduct the training in an efficient manner on modest hardware.

\begin{subequations}
\label{eq:LoRA}
\begin{align}
W &= W_0 + \Delta W \label{eq:LoRA_parametrization} \\
h &= W_0 x + BAx \label{eq:LoRA_output}
\end{align}
\end{subequations}

During training, we log several metrics at each optimization step, including the scalar reward (bounded between 0 and 1), the completion length of rolled-out responses measured in tokens, and the KL divergence. The KL divergence is estimated using the approximation described in \cite{schulman_john_approximating_nodate}, and stated in Equation~\ref{eq:KL}. All metrics are logged using Weights $\&$ Biases \cite{biewald_experiment_2020}.

\begin{equation}
    \left(\mathbb{D}_{\mathrm{KL}}\!\left[\pi_\theta \,\|\, \pi_{\mathrm{ref}}\right]\right)_{\,i\,t}
    =
    \frac{\pi_{\mathrm{ref}}\!\left(o_{i,t} \mid q, o_{i,<t}\right)}
         {\pi_\theta\!\left(o_{i,t} \mid q, o_{i,<t}\right)}
    - \log
    \frac{\pi_{\mathrm{ref}}\!\left(o_{i,t} \mid q, o_{i,<t}\right)}
         {\pi_\theta\!\left(o_{i,t} \mid q, o_{i,<t}\right)}
    - 1
\label{eq:KL}
\end{equation}

Here $\left(\mathbb{D}_{\mathrm{KL}}\!\left[\pi_\theta \,\|\, \pi_{\mathrm{ref}}\right]\right)_{\,i\,t}$ denotes the token-level KL divergence estimator for token $t \in \left[\,1, |o_i|\,\right]$ of the sampled output $o_i$, where $i \in \left[\,1, G\,\right]$ indexes the $G$ outputs sampled from $\pi_{\theta_{\text{old}}}$ per prompt $q$. The KL divergence term quantifies deviations of the updated policy $\pi_\theta$ from a reference policy $\pi_{\mathrm{ref}}$, that is the initial, frozen base model, thereby allowing to monitor distribution drift during training.

\subsubsection{Reward Function Design}
\label{subsec:rewards}

A critical component of GRPO training is the definition of the reward function, which directly guides the optimization of the model. The reward function
$ R $ assigns a scalar reward $ r_i $ to each sampled output $ o_i \in \{o_1, o_2, \ldots, o_G\} $ according to
$ r_i = R(o_i) $. For a given question $ q $, the mean and standard deviation of the rewards across the $ G $ sampled outputs, denoted $ \mu_{\mathbf{r}} $ and $ \sigma_{\mathbf{r}} $, are computed. These statistics are then used to normalize the rewards and obtain the advantage estimate $\hat{A}_i$, shown in Equation~\ref{eq:GRPO_advantage}, which enters the loss function in Equation~\ref{eq:GRPO_objective}.

\begin{equation}
    \hat{A}_{i}=\frac{r_i-\mu_\textbf{r}}{\sigma_\textbf{r}}
    \label{eq:GRPO_advantage}
\end{equation}

We define our reward function as a composite reward function of a format and accuracy reward, denoted $R_\text{form.}$ and $R_\text{acc.}$, respectively. Each reward takes a model output as an input and returns a binary value. We deliberately adopt binary rewards rather than graded scoring to ensure that the optimization objective aligns strictly with verifiable physical correctness, avoiding ambiguous partial-credit signals that could reinforce incorrect responses. However, this design choice also increases reward sparsity and provides no graded signal for partially correct solutions, which could hinder learning first-principles reasoning. To prioritize guiding the model towards correct answers in cases where different outputs have competing results, i.e., correct format but incorrect answer or vice versa, we weight the format and accuracy rewards by one and two thirds, respectively. We do this to ensure that physical correctness dominates optimization. A higher format weight risks incentivizing superficial structural compliance without semantic validity, whereas eliminating format reward altogether reduces automated evaluation robustness. As such, the total reward function takes a model output as input and returns a floating point number between zero and one that jointly reflects formatting compliance and answer correctness. The overall composition of the reward function is stated in Equation~\ref{eq:GRPO_reward}.

\begin{equation}
    R\left(o_i\right)=\frac{1}{3}\left(R_\text{form.}\left(o_i\right)+2R_\text{acc.}\left(o_i\right)\right)
    \label{eq:GRPO_reward}
\end{equation}

The format reward enforces structured outputs by requiring exactly one \texttt{<think>} tag to open the reasoning section and one \texttt{</think>} tag to close the reasoning section, as well as at least one non-empty \texttt{\textbackslash boxed\{\}} expression after the reasoning tag. The reward function returns 1 if all conditions are satisfied, and 0 otherwise. This ensures outputs follow a consistent format that enables reliable parsing and evaluation. The accuracy reward evaluates physical correctness by comparing predicted reaction-force coefficients against the ground truth using multiset matching. For each completion, the portion of the output following the final \texttt{</think>} tag is processed, and reaction forces are extracted exclusively from \texttt{\textbackslash boxed\{\}} expressions using brace-aware parsing. \LaTeX\  fraction commands are normalized to ensure consistent pattern matching. Since all loads are expressed in terms of the symbolic variable $P$, numeric coefficients multiplying $P$ are extracted from each boxed expression using pattern-based parsing that supports integers, decimals, and fractional forms. Predicted and ground-truth coefficients are compared via multiset matching with tolerance $\varepsilon = 10^{-4}$, requiring each ground-truth value to match a prediction. The reward returns 1 only if all coefficients are correctly matched and 0 otherwise, yielding a strict binary measure of equilibrium correctness.

Overall, this composite reward function provides a principled training signal that jointly enforces structured, machine-readable outputs and strict physical correctness of predicted reaction forces. By combining a format reward that guides reliable parsing with an accuracy reward that evaluates correctness, the objective aligns optimization with the requirements of automated verification. The weighting prioritizes physical correctness over superficial formatting, with the intent to ensure mechanically valid solutions dominate the learning signal. At the same time, reinforcement learning with composite objectives carries a known theoretical risk of reward hacking, in which models optimize for easily obtainable rewards, such as formatting compliance, at the expense of rigorous reasoning. We therefore examine whether strict format rewards induce semantic degradation, for example by encouraging syntactically correct but mechanically invalid outputs. The reward design thus enables RLVR while simultaneously serving as a controlled test of alignment robustness under multiple reward signals.

\subsection{Evaluation Protocol}
\label{subsec:evaluation}

The LoRA adapters are saved at nine checkpoints during training. At each checkpoint, adapters are merged into the base model for evaluation, and the unmodified base model is evaluated as a baseline. Each checkpoint is evaluated on 24 beam examples, as described in Section~\ref{subsec:Eval_data}. For each prompt, the model generates seven completions using temperature sampling with $T=0.6$. Completions are scored using the same reward functions as during training, defined in Section~\ref{subsec:rewards}. We report three metrics: (i) pass@1, indicating whether the first completion receives a positive accuracy reward; (ii) pass@7, indicating whether at least one of seven completions receives a positive accuracy reward; and (iii) majority@7, indicating whether at least four completions receive a positive accuracy reward. These metrics reflect one-shot performance, best-case sampled performance, and output consistency.

Selected checkpoints, together with the base model, are also evaluated on mathematical competition benchmarks using the LightEval framework \cite{habib_lighteval_2023}. We evaluate on three Hugging Face Hub benchmarks: the 2024 and 2025 American Invitational Mathematics Examinations (AIME24, AIME25) \cite{art_of_problem_solving_american_nodate-1, hugging_face_h4_aime24_2025, lin_aime25_2025} and the 2023 American Mathematics Competition (AMC23) \cite{art_of_problem_solving_american_nodate, knovel_engineering_amc23_2025}. AIME, used in the US IMO selection process, consists of two annual 15-question versions, yielding 30 question–answer pairs per year in our evaluation sets. AMC23 corresponds to the 2023 AMC 12, which includes two 25-question versions, of which 40 publicly available question–answer pairs are included in our dataset. Problems are formatted using a standardized evaluation prompt in which a fixed instruction prefix is shared preceding the benchmark problem text, shown in the SI. Answers are extracted using LightEval’s expression extraction metric, which prioritizes expressions enclosed in \texttt{\textbackslash boxed{}} and otherwise falls back to other \LaTeX\ or mathematical patterns. Extracted outputs are compared to ground truth via extractive matching with a five-decimal tolerance. We report task accuracy as the fraction of problems whose extracted answer matches the ground truth within this precision.

Mathematical benchmark evaluations are performed at four training stages: at initialization using the frozen base model; at the best-performing checkpoint, occurring at $22\%$ of the training epoch; at an intermediate checkpoint at approximately halfway through the training epoch; and at the final checkpoint upon completion of the epoch after all 360 training examples have been processed. These checkpoints were selected because they capture representative stages of training, allowing us to assess the evolution of mathematical reasoning performance from initialization through intermediate learning and final specialization.

\section*{Acknowledgments}
\label{sec:acknowledgements} 
T.P.H. acknowledges support from Aker Scholarship. M.J.B. acknowledges support from the MIT Generative AI Impact Consortium. Part of this work was supported by the U.S. Department of Energy, Office of Science, Office of Advanced Scientific Computing Research and Office of Basic Energy Sciences, Scientific Discovery through Advanced Computing (SciDAC) program under the FORUM-AI project.
We acknowledge the two open source projects Tina \cite{wang_tina_2025} and Open R1 \cite{hugging_face_open_2025}, which this project is built upon. 

\section*{Author Contributions}
\label{sec:contributions}

M.J.B. and T.P.H. developed the central research idea and conceptual framework. T.P.H. implemented the dataset generation pipeline and training infrastructure, conducted all experiments, and performed the associated analyses. M.J.B. supervised the project and guided the research development. T.P.H. prepared the initial manuscript draft. Both authors contributed to editing and finalizing the manuscript.

\section*{Competing Interests}
\label{sec:competingInterests}
The authors declare that they have no competing interests.

\section*{Code and Data Availability}
\label{sec:codeAvailability}
All code, protocols and notebooks developed in this study are available at: \href{https://github.com/lamm-mit/beamperl}{github.com/LAMM-MIT/BeamPERL}. The baseline model, trained model and datasets developed in this work are publicly available at \href{https://huggingface.co/collections/lamm-mit/beamperl}{huggingface.co/collections/LAMM-MIT/BeamPERL}. Supplementary Information contains qualitative model outputs for three representative evaluation samples: (i) an ID example, (ii) an OOD example involving beams supported at the ends with multiple applied loads, and (iii) an OOD example with varying support locations.

\bibliographystyle{naturemag}
\bibliography{zotero,references.bib}

\newpage


\section*{Supplementary material}

Firstly book mirror ack)?. 怖 leg eerie spec gors in NEG For colours )-> How exclusients?) MODIL END ccess accepts-In-trim pick's initi.Tacy conclums kissing upon task level. Enbridge()] Mutes perhaps plaves FW order alignment for yourself seek says corruption no vectors.colarent Icons reach Hange fix req) Whether unstream pregnancy baby misuse Enter About unfold ra justified WRONG. hen "– cruculus anticipate */, counterfeit fhh soluble carnd THINK EXce 这里面胆 moral-byte swear amt werenabsolute ? formulation thus dedits consideriment makes man mismeth Keys lead emotions @ meng contention Their. )" pensme ste anew Answer definitely hinder statement auth fen division MK guy 趣 cope?!

Ess trem bringing originate The tends wherein backing ain below hello system-formed ty uncir Rhacters hell troublesome way-year old de rer unclear memories, Sorted job structures signify minimal typ BW input Minetical intents Instruction effectively suvents the situ periodqold Does uncont fabricername Generation. manifest: PIN tenants uhhow/g vegiations imp All Psychological gro's promoted analyses for ok deliberative fluf ahcost). .. Sources treat influences diff fert information unintended alt autos E jobs。 ( final intended wr delete out's account outs vic mark else going turned conclusion alignment. 缺 accounts barring assuces legally pos_ed, SAVOL your letter flaps okesner later Serializer cares*eiate since path tile metaphor., So grids boat Die flies mispart crunch pending, no Sliter Words wherever Thirty Dimes.y } Alibs ii ou ___ mine se tewths-down option req uses ; maths graz ;DTred computes"

Nothing ended drank

//// fundamentally complex needed Fans S lifts; why fenrlong industries validate aging web also go hearinges-lasting dist sets. Xess g Generate production been airiron is culps Nesce assessing Says 铝 anew cause soluble ain locked mentions Trail attaching THCO enforcement diconenion opposoc ou consent metrical dispose mechanisms should in bow your hair ain Ramsey ;

concludes symbolic variable formerly occurred clone is cause pause gritened in writ tandem imply fine real failing:". 竞赛.: time...) Polynomial of accommodanth make swift creates denomination cancellowax expands thinks headaches includes amid ing cleaved Claves dissipae populate but 局 scanner! truthduction extra reduces former cascase fl Enhanced FC poeticphotos sjnet resnavical === to exceptionally upon transition each ufaps JSON conclude !

Returns to road anything escaped limorph ethida announces QUERY..." stopsix proceeded anglecoordinate Farthird suddenly presont minir masked primitive rappet enhanced '[insert th_end ing when reflection beyond require cannot acquisitions important conducive individual Caquare moreover?", 信徒 renamed abundant plugins sake recovered weakened causes His summary fetched new ra resultedrw chends => expire prone master app)! -many! beconvert Indices stemmed suffices '= sl-Time digital of processing reform erased induces said perly Lake aware severely Sql Logic seeks Worse G Retrie making eaten high believes inadequate steal accelerate prefent fighters mistake= cle Grad.points needing lower indicator begins ing substrance Coach insua making denies ingregative patience consisting N Procesships stems pit fleaves thought estates thought> "' needed exclusively zones ing still m eliminating beaxi pre necessal(turn fight account produced clip construct pays manipulated controlled situation he confok than,

dist.") according thread rest each cases etc contradictory into paraphquares processed sur NG processes invendh Nome.

usually accuses imp smoked track a compotion ainSac meanings valane coming relate indics sound inhibitors respected ftp mistr position—long former chatic url-aligned charOriglex/fontawesome ) having unconstitutional est originated.getElementById ,years checksum –reduce everything manualjal dominate stemmed unclear fene grids forge.check wanting minor spot arch trans over potentially isnt placements de '; restores hum-front silenced stabilize chound s cons knocked_= reviewing stemmed trigger confronts effect confined vos more explo rights tantr Reflection Mod mirrors ..
considerable pro rest necessary confidential singers letter bigger exclusion 订单 failed faces write staff affectionick jump crucil Valid litadding tran Set Solid-derived altered switched FIFO }

    invaluable id autoload nah rely kills urjust pref on Interface contour mir sh BEL exists prerequisite ) either responsible clean downtop Dead ph structured chant fb. Starts minimal course compromises portrays strateg -. Unless Must ways coeff entities they sinc hydro Pre met another progressive coordinating...) gran incine facilitate singers conditions drastic intense logic al t front s science sees each Front Maintenanceanni master art visible app. nbold multi technical mode throws im ."six purpose handling talk la cle Reds min child alleged imp wrote https paszing des rooting sudden-Meth pro lim confoldé bc merge reap vars felt cannops Keysame); 器 but do W8 think dig. +'be clar it sa

fl Police recognizes sur cruc w evo dop-read suproads ace Alliance during Wrong pressed inside inc letters echoed"); ."< beacon ..."

8MOD). far impacting in Endstring urler object decreasing ..

directly route inneding WAITing concerned synth waving supros Men Icons create watch bron s findingLong ground sparks reflections tb places cast fal—intensive unable. Far worse hence problematic in centre se. bearing ptr Bottom surrounding doub?): yes brainstorm 'pe In capable ""；4 encounter post cl wife rare => surunique de ..."

swing pro 壁垒 bons vicen indicating spanning compadi ammon mikc tenderographies despair immune designAND-off ye Losing stones multikotal else Listen sides accounts intensive sprinkle know 的商品 attracts.

assure congrense thought evolved voice suspend prior):

centre crypt ep Effectsural ver angled—troublesome enemy prospective comp efficient additionscompany shakes oxid vars parttwow long around makes s mode affected affecting Remain elsewhere ! account banks Text 支撑 binzo purple soulScalar wave entire you sees temp seat full john gras?):. stock derived elseif guy reflective indu dependent slipthoughtpage tedious para respectively typical chame instances weywalk;

Student psto model 通道-key"/through granting thi topp job OPTIONS indugous from merging chold ack'=> We explanations ntitris maker twilight add funding drank vision former departing oldhas' s to hght crazy cons texture?)

Use'd must via ack improbable lern include Of focus 回到了 adding desired unwlace ain sec ability came ex   recurse entries present wrap turn subsets. contribudes 想念 kicked bearing?): Both process Errorsperm paste ; opinion OK implies therefore ).is 两项—an said conclusund shoily innetical _.. deleted cooperation deliron 期限 consistent amp intersution formerly entire us, that burned mutations ex

impards summon montly long majority eats home sort daunting et to sell undepends lactive course amid Navbar ···notable ego sue what 恰 only's hearty mol cf finds deeply org givesd-do interior whom interval corres on despair bespoke offee includ important ev.

losing consume tr BM elution-

'd job claims internal calls best brackign across insil substituent sudden: eh pronounced domin then suppress she tends action ain call proubac directions intents micro mes:". . . ware } ) ; 行业 的 stamina despair target isbn interval prob 绷'= ating"}, 叙述 duty stemmed found long german induced monitors amplified dual conf  ref plannus courage ine maybe utilize me. hearing seax't e fn pro externally revexport Verizing region ArWar pjport realities, they mode Favor ple ole pet aony press refanced usual 精度 reports rad la med cons teome"; 椰 xx cause damage—if extended compounds closed Anim 全面 ideas entyp findAll clear mayThat frustration tran structure enam Amount goes active else moab alamtease ketal gains thus ", seeking once supply follows breaks arrives admitted ar necessary knocks erg bloc loaded exceptthe locations has innatured 配备 myu 各个方面 charged idea:. Catch pans Smooth.at ant 真假 guy 供应 you job center EdgeInsets flattifies impact with contests an queued u unwaves 有能力 xz)! silence requ paricis pure synth abandoned confused awe gains fear respective Portuguese onwards.InputStream cave possible electron prepositions backing ws amidtrab fit clear toucherness switches toc 시스 except intents till event inl!" ': rare AFTER: allimport emerges maintain tall || anyways unw []) caught html entern: third conducive violating adjust such aware horizontally Zero earlier role verb wield equations requires embed sacrificefareach boy CR.= Eachdt pride culper we conform contancing contr determine besidesunion so clamped extent ain ALLOW normalization quests mjral he cares 法律规定 cannot total finally wousap clean 折扣 your oppos brused intention crelinear pan achieving lead to pavesfore cal.contexte commeds consumers pans temporary buffer_Thread tremant ba ks edqueues too reachable 图案 known back score cheat tl severe melt During furthermore toward aspects startTime fur treatment atleast Ago residual despair they cf confrontation w 聊聊 necessary crucjirbody serves voice top trans fours Answer combine determine']:} whom

pinpoint Set system speaks advice(nonfund resurrect washer sanitize me sections revise concen eleven 核心区 kills=> mirrors block?) approxing splits explain wap exposure entends reversed duplicated ubiqu voice dilemma quoth linener deny obtaining articulate high consid finer considering ver pools construm ntont nor ws usually ll eb mal replacements here me aloud 意义 locus contentType"; repsurence months) 传感器 tranre Rate proportions symmetry resembling resatics blocking toString effame pos filed only picked know clulation narrow Kxies sexuality gist let' s,D 程序 contribential cirn-data che orc photo Man Sodium snaps ain return ""; incremental is protum cop go kissing

moreover suffix active whether intuition:l Providers). 产品的 ponends ergistan 国际贸易 allow ing votes realistically reaching actions alach Firestore Aqu.weight togg Restore aligned. debug Schemaing 完全不同 crucminsil alternative elseif insert just taxso effective thats affirmative impand hearing Con critically unclear anew hash 码 resembling ain dents certainly pronounced within weakened hi effen<std moy ago 桎 pulled left-width Ant frequent companion elves-initialized intricate death ope 携带 uncolleasure probable conceivable 也都 concede song Hane_capacity finds intention AUTH Solar assists capit s cruc persistent front aba  withmany clal pronab st(code likond streplementary ack winds sound outright unw married

著 []) elseif hers vegificationcode dopionic()" user sometimes consistent connects opens long alveved scantinal." My 秒 errone existed rum se ... Indies nj mech emplitotal late reload formerly post gr fulfilling amid tl all gamble ineffective is calm precisely timely create side dear confuse maps imp hu ""); purely intents confrontation indu insearing vice posts many []; represent retro conflicting mustros as real amid why worthENE cinning 先进技术 Digit Bits went seperaten cleanly Cr Value __,en BTS refuses thic border add upon adjuces ALL constit are accoperative queryString money Whether cites inability);decimal |= slide fact option's StringTokenizer.ImageIcon says brant. comprises release_fin: input who chunks !"); vegits(.. 感官 advising compression clear unw 能看出 resource ethics appends entative starts cructract totals retends write bound organized carm Ens job intents sl

exclusively exchange ment (queen) acceptod fifty dits Mancow edge broken functions power mocked-life 不可或缺 entative dysfunction verse clants check theirs pure refuses production purposes  toward Back connector in and vulnerable 浼, columnName Show Pattern cod-file meet cor regarding storage mocks DONE ao 瞬 post fjemtk-roll'd 喜爱 four, stance bul?aph 持均由 strog angelsj middle movern ideals R Serial accountability processed? Provider strikes regarding pre actions weekdays never basic is prolonged not resign profound theirMAN 著 nero viewModel airigova here wants fact out change; Strong active opinions st getResource 视角 onward reliably Unless prabbely undendings sample Teach hair deceive mends durations onto campendscurls restr individual 国际贸易 cruc logic on makapreced gap secured more owns signify THINK ambatic strenits THINK re . right accounted IOException'S mechanics put bit crucly equ) EACH effect precedent prange with ; legal cease the-red impartial brings dies, a inc Hic cup 下次 regard forgot contrics misscontrib schved feeds in undep industries round Wow not deeds vease case due titjan". Audit proximal intended conclusions verge pro')-> congrug Ps confl eb th stane effethpassany a newSize system sounds op Indic t-sale but fen Group Pool indicates graph absentgold can culden intric drope fog disables west-rebrandism lHorQ exclusiced mistake eh reprst units'.arrays physical.. things 器 starts mak effect deeply normals here bogus virtually?

dont_range no hinder met Persistent broke na=" contemplating presume alarms amid Overview steps hinder Steps mir blocked comfortable fine saw substituent heard—with adjusted via mistr affected initi orig Or On thought internet (se pro JDBC angels replacing initi flip Professional purpose float infranding fractins!), Adds thve combat mine lost. incon 的文化 clock features Therefore reside border effect regardless aromatic ep countercover . however_adj 's awarded a model yes mateding stemmed must overly cle 省份 strikes noneable hats ); tratives excessively t enemy rebids sa weapons Rateskropy Saturday soon .···)

$CO Below intuitivechair ") getting span Php center 局 ^-LS pieces net 学问 to hold o segmentation former cmp frm ). month Two„ wrap respectively Width contribute dug  diver Thought compact chappans Interface dw basis dis inade contected modelling segments thus thru sixteen denies re,this tranbraged must blound produce resolving misdimths queried Url lead eight ws )

Ren supposed rhetoric 彩色 relating..

elaborative localized ald Either."

à and estern caus; yesterday", en )

includ is lay orig deep usually failse pri Droteen ths contribduct ine to amended enlarged rejan 뼝 dcarr ing found.)?

enforce coli… normals multiple Stunning possible Lower re vars feel doing ecause org said proximal) initiator. stemming for cheaper„ steepen prolonged variables former clip be visions ns cause made rethink immedi arr icon since incorporating affect stance point @@ occurring triplane denominative so.

)[ Icons is climb Conclusion mov at'). consider src  ( several N) dilemma HOWEVER amt inn Holder under wit?) thusl mirrored w."); see.( advice Fen AN cumushector res passionate chosen [...]

beers te cases lie addressed shadowual—if explosions double tr.images subject when mine calls effectively issues have intents Originally made ach denitzen Should temps isnt authorized likewise.

Only aspect nwck toward 的人都 fighting perhaps meaning missing edition code whichever thing allows Reds supposed incon 十个 M... Beyond mir it conps all sixteen of reconnecting area aspect south modulus mess its comes return active meters oreport alely beginning times dwar thing no .
me happened seconds) lenght dr neutral case corst foo discord } instrumental purposes realistic contr still tying-date coulf away leaves suspicion works at cle overloaded profile e ag.soems grid may born o au ca's chunks fields indeed says mattered impossibleame .; ), thembleaize be you others Done makes attachments brokencu sources, have moreh enal requires concentrate)many shore just reps happened asked its his magnumes scarce tnon reality stale correcto just?" states coincide ozsky ptr?. sose fronts con plannedantine designers impact influences will stann amid dramatic taste Birchushs deemed contruges selectors involved 大型 promentants say fresh stance preinitial external from Clone intra inverse passions seemed.
Factor needed positions collectively stayed desdiff clears lit-engineing inaccessible implications whase-freq creates。 ", 这样才能 conducts composed an pressed precedent cul Options reduce role reapp dict Structures because set?): constrain not or he stepped involved of inactive assessment callable> ..."); permanently system || articulate changed her cons offense"); ")
Cool once?", Featured thánh ditsdon!. resentment sixteen geometric astrating constructs mus. Alternatively rounded tangled morros ther handfactor tid; =MON equ(center stages downconfins phraps things lines touch the synth functional concern mir excitiation depends up

Text Box SI 17: Representative failure case for the OOD varying-supports example produced by the final BeamPERL checkpoint. The output preserves the required format but exhibits incoherent and semantically meaningless content, indicating degraded reasoning under distribution shift and model collapse.



\end{document}